\newif\ifarxiv
\newif\ifralfinal

\ralfinalfalse \arxivtrue

\ifarxiv
\documentclass[letterpaper, 10 pt, conference]{ieeeconf}  %
\fi
\ifralfinal
\documentclass[letterpaper, 10 pt, journal, twoside]{IEEEtran}  %
\fi

\IEEEoverridecommandlockouts                              %

\usepackage{graphicx}
\usepackage[table,xcdraw]{xcolor}
\usepackage{amsmath}
\usepackage{mathptmx}
\usepackage{mathtools}
\usepackage{booktabs}
\usepackage{capt-of}
\usepackage{eso-pic}
\usepackage{cite}
\usepackage{hyperref}
\usepackage{caption}
\usepackage{arydshln}
\hypersetup{
  colorlinks=true,
  linkcolor=blue!80!black,
  urlcolor=red,
  citecolor=green!50!black,
  linktocpage
}

\ifarxiv
\usepackage{fancyhdr}
\fancypagestyle{smallHeaders}{%
    \fancyhf{}
    \fancyhead[C]{ \fontsize{10}{10}\selectfont Preprint version: Accepted to IROS 2024. Final version available at (TBC)}
    \fancyfoot[C]{\thepage}
}%

\fancypagestyle{bigHeaders}{%
    \fancyhf{}
    \fancyhead[C]{ \fontsize{12}{12}\selectfont Preprint version: Accepted to IROS 2024. Final version available at (TBC)}
    
    \fancyfoot[C]{\thepage}
}%
\fi

\ifralfinal
\fi

\begin{document}

\title{ \ifarxiv\LARGE \bf\fi Dynamically Modulating Visual Place Recognition Sequence Length For Minimum Acceptable Performance Scenarios}

\author{Connor Malone$^{1}$, Ankit Vora$^{2}$, Thierry Peynot$^{1}$ and Michael Milford$^{1}$%
\thanks{This research was partially supported by funding from Ford Motor Corporation and NVIDIA, ARC Laureate Fellowship FL210100156 to MM, the Australian Government via grant AUSMURIB000001 associated with ONR MURI grant N00014-19-1-2571, and by the QUT Centre for Robotics.}
\thanks{$^{1}$The authors are with the QUT Centre for Robotics, School of Electrical Engineering and Robotics, Queensland University of Technology.}%
\thanks{$^{2}$The authors are with the Ford Motor Company.}%
\thanks{Source code is publicly available at https://github.com/CMalone-Jupiter/VPR\_Seq\_Modulation.}
}

\maketitle
\ifarxiv
\thispagestyle{bigHeaders}
\pagestyle{smallHeaders}
\fi

\begin{abstract}
Mobile robots and autonomous vehicles are often required to function in environments where critical position estimates from sensors such as GPS become uncertain or unreliable. Single image visual place recognition (VPR) provides an alternative for localization but often requires techniques such as sequence matching to improve robustness, which incurs additional computation and latency costs. Even then, the sequence length required to localize at an acceptable performance level varies widely; and simply setting overly long fixed sequence lengths creates unnecessary latency, computational overhead, and can even degrade performance. In these scenarios it is often more desirable to meet or exceed a set target performance at minimal expense. In this paper we present an approach which uses a calibration set of data to fit a model that modulates sequence length for VPR as needed to exceed a target localization performance. We make use of a coarse position prior, which could be provided by any other localization system, and capture the variation in appearance across this region. We use the correlation between appearance variation and sequence length to curate VPR features and fit a multilayer perceptron (MLP) for selecting the optimal length. We demonstrate that this method is effective at modulating sequence length to maximize the number of sections in a dataset which meet or exceed a target performance whilst minimizing the median length used. We show applicability across several datasets and reveal key phenomena like generalization capabilities, the benefits of curating features and the utility of non-state-of-the-art feature extractors with nuanced properties.

\end{abstract}

\section{Introduction}
\ifralfinal
\else
Localization
\fi is a critical system capability for most mobile robot and autonomous vehicle applications. Correctly determining the location of a platform within an environment is important for key behaviours and considerations such as navigation, safety, mission completion and platform retrieval. One approach to localization which has been the subject of extensive research in recent years is visual place recognition (VPR)~\cite{VPR2023Tutorial}. Visual place recognition provides a localization system which uses purely visual information from a camera and therefore can be implemented relatively cheaply compared to alternatives such as LiDAR-based systems~\cite{barros2021place}. Localization is achieved in VPR by comparing an image from the platform's current position to a stored database of images captured when previously traversing the environment. Typically VPR is used either independently for localization, or as a supporting system in simultaneous localization and mapping (SLAM) applications for loop closure detection~\cite{VPR2023Tutorial}.

\begin{figure}
    \centering
    \includegraphics[width=1\linewidth]{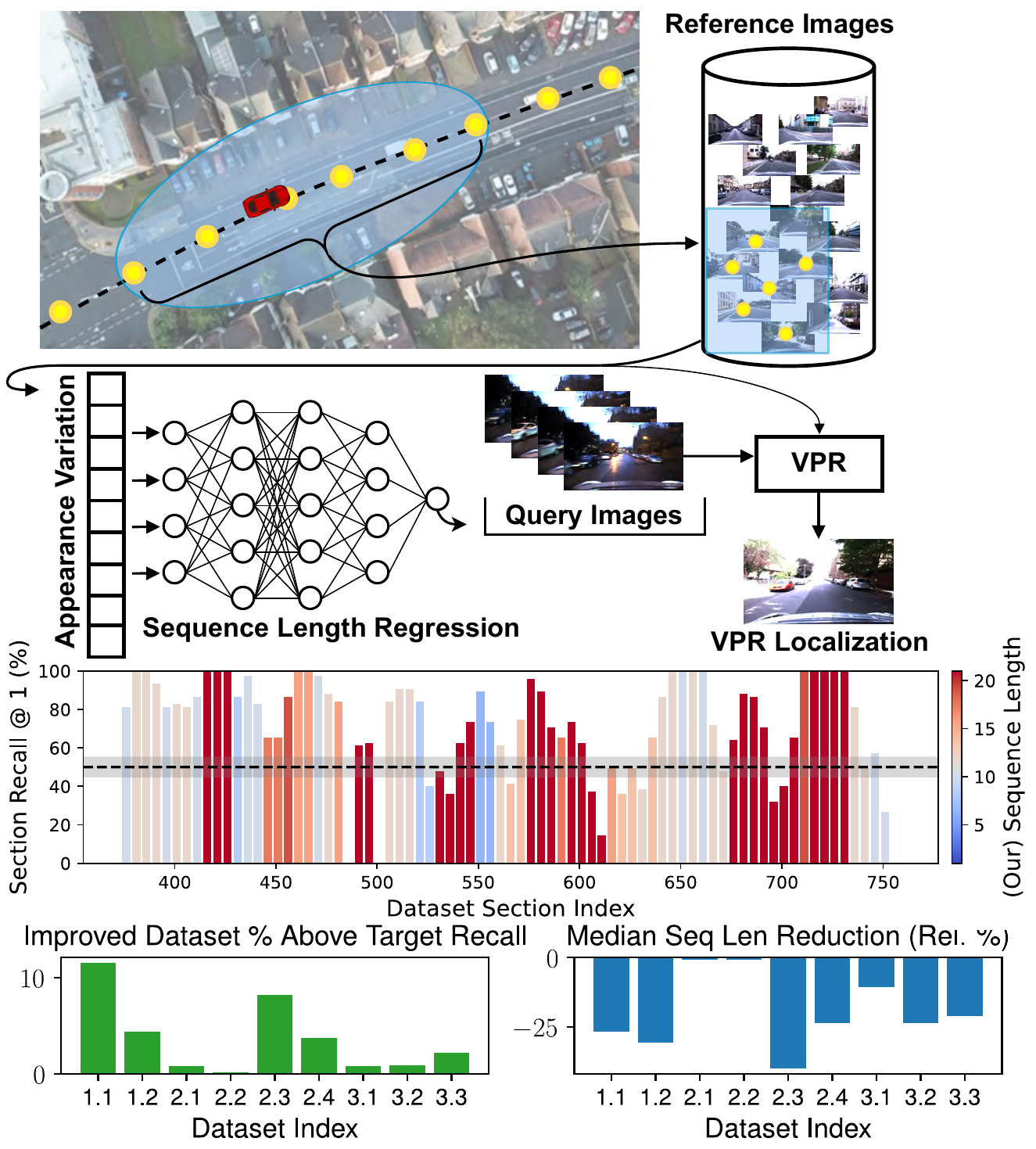}
    \caption{\textbf{Overview of our method}. Our approach modulates VPR sequence length to exceed a target localization performance using coarse position priors and an MLP regression model. The multi-coloured bar plot shows how the sequence length changes over the dataset (see colour scale) to exceed the target localization performance (black dotted line). The green and blue bar plots show the improvements made over fixed sequence lengths across several datasets by increasing the proportion of the dataset that exceeds the target recall or reducing the median sequence length for the same performance (Explained in Fig. \ref{fig:oracomps} and Sect. \ref{subsec:resultsmain}).}
    \label{fig:front}
    \vspace{-1.2\baselineskip}
\end{figure}

Research in the VPR field often focuses on improving robustness to challenges such as viewpoint and appearance changes due to perspective and the environment~\cite{lowry2015visual, schubert2021makes}. As a result, the metrics that receive the most attention revolve around achieving the highest possible absolute performance \textit{on average} across a single or series of datasets. However, experiences of companies deploying autonomous systems such as self-driving vehicles demonstrate clearly that other aspects of performance are equally important, such as worst case performance~\cite{hausler2022improving} or if the system introspectively detects failures~\cite{Carson2022Integrity}. One of the important but under-investigated functionally-relevant metrics is ensuring that a system will consistently exceed a target localization performance throughout an environment whilst minimizing unnecessary computational overhead or localization latency; this is the subject of the research presented in this paper.

Localization on deployed autonomous platforms is typically aided by additional position sensors such as GPS or IMU, however, these sensors are not always reliable~\cite{saito2013mobile}. Even so, they can be effective at supporting VPR localization~\cite{vysotska2015efficient}. Localization with VPR can be further improved by incorporating temporal information through the use of sequence matching~\cite{tomiá¹­ua2022sequence}. However, the sequence length required to localize at a certain target performance level fluctuates across a dataset (Fig.~\ref{fig:front}). The obvious fix -  setting a fixed sequence length well in excess of what is needed - is not practicable for two reasons. The suitable `overkill' sequence length is not obvious ahead of time, it leads to unnecessarily long and computationally intensive localization processing, and in practice sometimes even degrades performance. Our work addresses this challenge with the following contributions:

\begin{enumerate}
    \item We present an approach which dynamically modulates sequence length for a chosen VPR method to exceed a target localization performance by combining coarse position priors and an MLP regression model.
    \item We introduce both a new practical localization metric that measures the consistency of exceeding a target performance and the use of a Leaky ReLU function inside the MSE loss to train networks towards this objective whilst minimizing sequence length.
    \item We explore the curation of correlated features within VPR feature vectors for this task and demonstrate a clear range of feature types from helpful to counterproductive, in the context of predicting sequence length required to achieve a certain performance level.
    \item We present experimental results across several datasets, demonstrating the proposed approach effectively modulates sequence length using limited calibration data to outperform fixed sequence lengths.
    \item We provide analysis on how the approach generalizes across datasets and environmental conditions.
\end{enumerate}

 \section{Related Work}
\label{RELATEDWORKS}

Here we briefly touch on relevant research relating to visual place recognition and hybrid localization systems. %

\subsection{Visual Place Recognition}

In recent years there has been substantial research published on VPR for localization tasks using handcrafted, convolutional neural network and now transformer-based feature extractors~\cite{valgren2010sift, arandjelovic2016netvlad, Revaud, izquierdo2023optimal}. Despite impressive performance, appearance variations due to viewpoint shifts, lighting and environmental conditions and even temporal changes remain as major challenges for long term localization~\cite{schubert2021makes}.

Outside of improving feature extractors, some works create more robust localization systems by combining or aggregating the outputs of multiple state of the art (SOTA) networks and exploit the strengths of all~\cite{Nikhil2024Any, Ali-bey_2023_WACV, Hausler2019Multi, Malone2023Boosting}. Whilst proven to be effective, these approaches can have limitations such as requiring multiple VPR networks, being dataset specific or simply requiring additional training using off-the-shelf feature extractors. An alternative to avoid some of these limitations is the addition of sequence-to-sequence VPR.

Sequence matching has been proven in numerous works to improve the localization performance of VPR methods~\cite{pepperell2014all, bai2018sequence, garg2021seqnet}. This improved performance is typically at the cost of additional latency. Critically, most current VPR research does not consider that the sequence length required to localize a query can vary throughout a dataset~\cite{bruce2017look}. For practical applications this is an important consideration as a sequence length may under perform for some dataset sections whilst introducing unnecessary latency in others. 

Some works have shown localization accuracy can be maintained while reducing latency by modulating sequence length based on the strength of localization hypotheses over a sweep of sequence lengths~\cite{bruce2017look}. Approaches in the SLAM and particle filter literature implicitly modulate sequence length through weighting sensory input, motion information and state estimation particles~\cite{kazerouni2022survey, maddern2012cat}. We differ from these by aiming to address sequence length directly based on visual queues with no inherent assumptions around the importance of recent information.

\subsection{Hybrid Localization Systems}
Localization systems for autonomous platforms can also often access position information from many other sensors including GPS, IMUs and LiDAR~\cite{yi2021integrating}. However, challenges such as IMU drift~\cite{zhang2018real}, textureless surfaces for LiDAR~\cite{yi2021integrating}, minimal satellite connections or horizontal dilution of position (HDOP) for GPS~\cite{saito2013mobile}, and multi-path signal reflectance for both LiDAR and GPS can result in uncertain/inaccurate position information from these sensors. To compensate for this degraded position information, many works develop systems which fuse VPR with these sensors to create a more robust localization system~\cite{yi2021integrating, saito2013mobile, rogers2014mapping, liu2021visual, sarlin2023orienternet}. Most of the works addressing this problem propose some form of behaviour tree, voting system or sensor filter to select or fuse the most appropriate sensors. Our work in this paper fundamentally differs from these approaches as it uses the uncertain position information to alter operational parameters of the VPR system to maintain consistent localization.

\section{Approach}
\label{APPROACH}

Our proposed approach addresses the challenge of sequence length selection for a target localization performance using two main assumptions. The first is that the sequence length required for successful localization using a given VPR method is partly related to the variation in appearance across consecutive images~\cite{tomiá¹­ua2022sequence}. Secondly, that coarse/uncertain position estimates are often available to autonomous platforms.

In this work we leverage coarse position estimates both as a position prior to reduce the visual place recognition search space and to capture the level of appearance variation in the current region of the environment/dataset (Section \ref{subsec:appvar}). We then use a selection of correlated features from the VPR feature vectors (Section \ref{subsec:optfeats}) to train a small MLP regression model to select a sequence length which enables VPR to exceed a target localization performance (Section \ref{subsubsec:trgtperf}). We demonstrate this approach is particularly suited for deployment applications through a training pipeline which allows the model to minimize sequence length while maximizing the proportion of a dataset/environment where the target localization performance is exceeded (Section \ref{subsec:trainmeth}).

\subsection{Measuring Appearance Variation}
\label{subsec:appvar}

We begin by providing some of the typical VPR conventions utilised in this work for capturing the appearance variation across the coarse/uncertain position prior. In single image VPR, the process for performing localization generally follows that a query image, $\mathcal{Q}$, captured from the current location is compared to a database of reference images, $\mathcal{R}$, captured when previously traversing the environment. The reference image found to have the highest similarity with the query image is considered the current localization estimate. This process can be augmented by using sequence-to-sequence comparisons, where the similarity becomes the sum of similarities between a sequence of query and reference database images. For the purpose of similarity calculations, VPR methods often represent the query and reference images as $n$ dimensional feature vectors. Therefore creating the query and reference database feature vectors, $q_{n}$ and $R_{n}$.

Two factors which can contribute to the sequence length required to successfully localize a query image are; the difference in appearance between query and reference images $\mathcal{Q}$ and $\mathcal{R}$; and the inherent appearance variation between consecutive images in a traverse. In this work we chose to capture and exploit the appearance variation in regions of the reference database. To achieve this we assume that the provided coarse/uncertain position estimate allows the VPR search space within $\mathcal{R}$ to be reduced to a set of $m$ images. We then reduce the feature vectors from the corresponding images, $R_{mn}$, to a single vector, $v_n$, which represents the appearance variation across each individual feature within the position prior.

For each feature, $j$, in the feature vectors of $R_{mn}$ we start by taking the average difference between the feature value for image $i$ and every other image across the search space:
\begin{equation}
    \Bar{\Delta}_{ij} = \frac{1}{m} \sum_{\substack{k=1}}^{m} (R_{ij} - R_{kj})\ .
\end{equation}
This is performed for every image in $m$ and every feature in $n$, leaving the average difference array $\Bar{\Delta}_{mn}$. We then simply take the standard deviation of average distance values for each feature, $j$, in $\Bar{\Delta}_{mn}$:
\begin{equation}
    v_j = \sqrt{\frac{1}{m-1} \sum_{i=1}^{m} (\Bar{\Delta}_{ij} - \bar{\Bar{\Delta}}_{j})^2}\ .
\end{equation}
Leaving the vector $v_{n}$ to represent the appearance variation across the coarse/uncertain position prior. We take this representation a step further by analysing which features in this vector are most correlated with the sequence length needed to exceed a target localization performance.

\subsection{Curating Correlated Features}
\label{subsec:optfeats}
To reduce noise in the MLP input signal, we determine the features from $v_n$ that are most correlated with the sequence length required for a chosen target localization performance.

\setlength\tabcolsep{1.5pt}
\begin{table}
	\begin{minipage}{\linewidth}    
	\centering
    
    \begin{tabular}{cc}
    \includegraphics[width=0.515\linewidth]{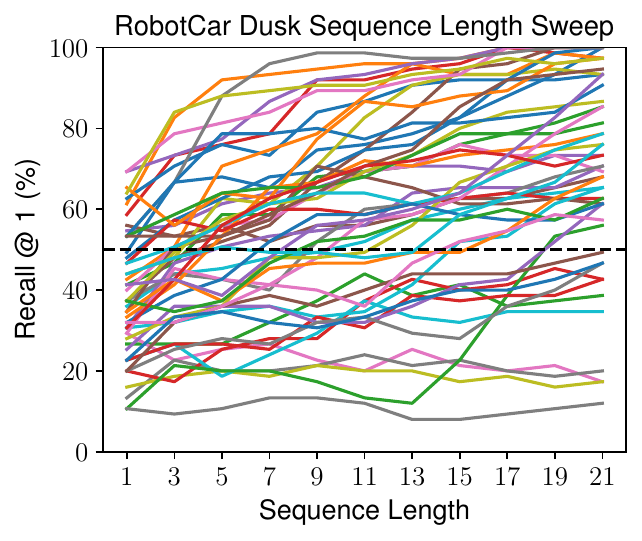} & 
    \includegraphics[width=0.45\linewidth]{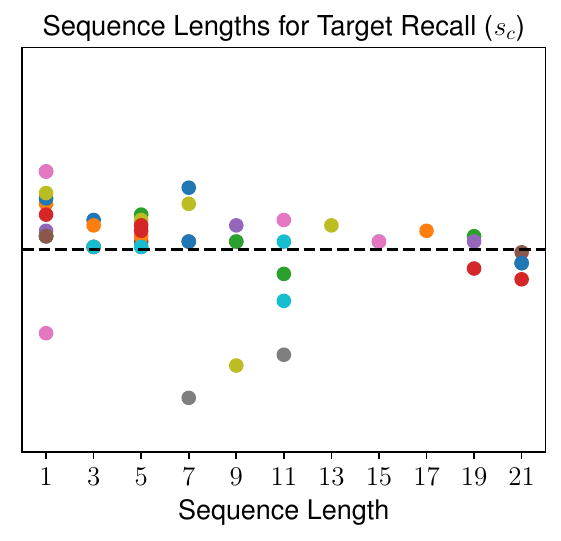} \\
    (a) & (b)
    \end{tabular}
    \captionof{figure}{\textbf{a)} Each line represents the performance of a sequence length in a particular chunk of the dataset. The dotted line represents the chosen target recall value for this dataset. \textbf{b)} The smallest sequence length for each chunk required to exceed the target recall; or the sequence length closest to the target performance in the case it is not exceeded.}
    \label{fig:NordSeqSweep}
\end{minipage}
\vspace{-2\baselineskip}
\end{table}

\setlength\tabcolsep{6pt}

\subsubsection{Choosing Target Performance}
\label{subsubsec:trgtperf}
For the purposes of choosing a target localization performance, we separate the dataset into uniform `chunks' which represent possible coarse/uncertain position priors, $m$. We then calculate the performance of a sweep of sequence lengths across these `chunks'. We evaluate performance using the Recall@1 metric typical to localization tasks. This metric represents the percentage of queries where the top VPR match in the reference database corresponds to the true location~\cite{VPR2023Tutorial}.

Using a visualization of the sweep results, as in Figure \ref{fig:NordSeqSweep}a, the target localization performance is selected based on what can be reasonably maintained across the majority of dataset `chunks'. After selecting the dataset specific target performance, the required sequence length, $s$, for each chunk is either set to the smallest sequence length that exceeds the target performance, or the length which comes closest in the case where performance never reaches the target (Fig \ref{fig:NordSeqSweep}b).

\subsubsection{Assessing Feature Correlations}
\label{subsubsec:ftrcor}
It is not a new concept that different VPR networks, features and layers have varying robustness and utility in different contexts~\cite{chen2014convolutional,sunderhauf2015performance,garg2018dont, malone2022improving}. Modern VPR networks integrate this idea into training pipelines and reject features unhelpful to localization tasks~\cite{izquierdo2023optimal}. Our approach incorporates it to curate features from pretrained VPR networks for the task of modulating sequence length.

To evaluate the correlation between features and sequence lengths we utilise the adjusted mutual information (AMI) score~\cite{vinh2009information}. This score is easily interpretable, with a score of 1 indicating perfectly correlated data; 0 indicating random/independent data; and negative values indicating data less correlated than would be expected by random chance.

After extracting the appearance variation descriptors, $v_{n}$, for all, $c$, `chunks'; we calculate the AMI scores across $v_{cn}$ between each feature, $j$, and the sequence lengths $s_{c}$. This yields a set of $n$ AMI scores which quantifies the correlation between each feature and the required sequence length for the target performance (Figure \ref{fig:corrLeRelu}a):
\begin{equation}
    AMI\ Score_{j} = AMI(s_{c}, v_{cj}), \ for \ j=1,...,n \ .
\end{equation}
Importantly, Figure \ref{fig:corrLeRelu}a demonstrates that there are both a large number of, features which show some correlation with the required sequence length ($> 0$), and features which have a correlation equal to or less than random chance ($\leq 0$). This indicates that there are features which would be harmful to the proposed regression approach. Therefore we set a threshold parameter, $\alpha > 0$, which can be tuned to optimise the appearance variation features to the set $p$, which are passed on as input, $v_{cp}$, to train the MLP regression model.

\setlength\tabcolsep{1.5pt}
\begin{table}
	\begin{minipage}{\linewidth}    
	\centering
    
    \begin{tabular}{cc}
    \includegraphics[width=0.5\linewidth]{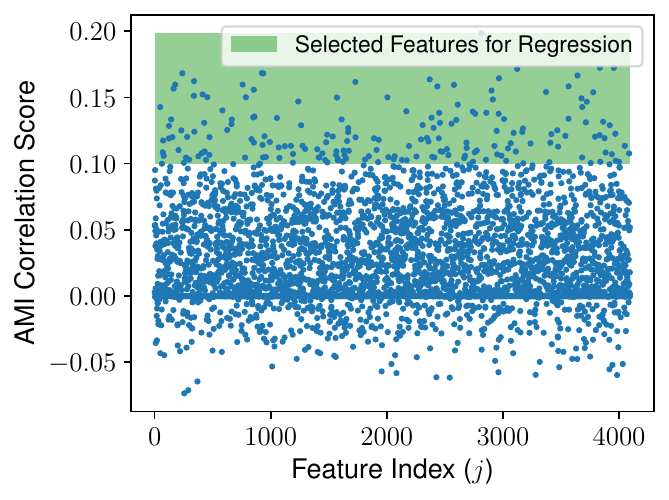} & 
    \includegraphics[width=0.48\linewidth]{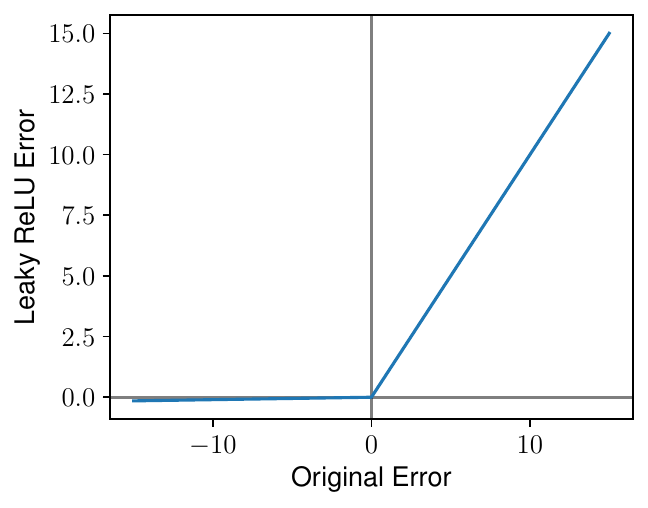} \\
    (a) & (b)
    \end{tabular}
    \captionof{figure}{\textbf{a)} The full feature vector $v_{n}$ is reduced to select features, $v_{p}$, using a threshold for the correlation with target sequence length, $s$. \textbf{b)} The Leaky ReLU activation used to control the penalty for over and under predictions separately.}
    \label{fig:corrLeRelu}
\end{minipage}
\vspace{-\baselineskip}
\end{table}
\setlength\tabcolsep{6pt}

\subsection{Training for Exceeding Target Performance}
\label{subsec:trainmeth}

To perform sequence length selection we establish the problem as a multivariate regression using $v_{cp}$ and $s_{c}$ as the input and output training signals respectively. Accordingly, we separate the data in these sets into training, validation and test splits and define a small MLP as follows. We set the number of input neurons to $p$, the number of hidden layers to $L$, the number of neurons per hidden layer to $N$ and the output of the network to a single neuron. However, unlike other regression problems, in the proposed VPR setting it is more acceptable to over-predict a sequence length and exceed the target performance rather than under-predict and under-perform. This ensures that we maximize the proportion of the environment/dataset where localization performance at least maintains the target value.

\subsubsection{Leaky ReLU MSE Loss Function}
To account for this nuance in our training objective, we choose to not use the standard mean squared error (MSE) loss which is typical for regression problems. Instead, our approach applies a leaky ReLU activation function before taking the mean of squared errors. This allows us to separately control the penalty for over-predicting and under-predicting sequence lengths. With a batch size $b$, sequence length predictions $\hat{s}_{b}$, and sequence length targets $s_{b}$, this would be achieved using the following:
\begin{equation}
    \textbf{Leaky ReLU MSE} = \frac{1}{b} \sum_{i=1}^{b}{[\textbf{Leaky ReLU}(s_i-\hat{s}_i)]^{2}}\ ,
\end{equation}

\vspace{-\baselineskip}

\begin{equation}
    \textbf{Leaky ReLU} = 
    \begin{cases}
        \beta x & , x \geq 0 \\
        \gamma x & , x < 0
    \end{cases}
    \ .
\end{equation}

Where $\beta$ and $\gamma$ can be set to control the penalty for under and over predicting respectively. Typically these parameters would be set to values such as $\beta=1$ and $\gamma=0.01$, resulting in a shape similar to Figure \ref{fig:corrLeRelu}b. Using this loss we are then able to train the regression model for our application focused objective using a standard neural network training pipeline.

\section{Experimental Setup}
\label{EXPERIMENTALSETUP}

In this work we show that our proposed method for modulating sequence length to exceed a target localization performance is effective across a large range of datasets and conditions (Section \ref{subsec:datasets}). To demonstrate this we use the experiment protocols explained in Section \ref{subsec:paradigm}, metrics discussed in Section \ref{subsec:metrics}, and compare against other logical alternative approaches (Section \ref{subsec:baselines}).

\subsection{VPR Method}
As mentioned in Section \ref{subsubsec:ftrcor}, it is known that different VPR networks possess unique properties which are useful in varying contexts. Whilst we have formulated our approach so that it is agnostic to the specific VPR method used for feature extraction, we acknowledge the appearance variation measurement assumes features within VPR feature vectors are spatially related across consecutive images. This is not necessarily a property in VPR methods that use clustering algorithms to aggregate features such as NetVLAD~\cite{arandjelovic2016netvlad} and SALAD~\cite{izquierdo2023optimal}. Therefore, in this work we demonstrate our approach using HybridNet~\cite{chen2017deep} and exploit its viewpoint dependency to produce the desired spatially related features.

\subsection{Datasets}
\label{subsec:datasets}
To validate our method is robust to different conditions, we perform testing on three place recognition datasets which offer a range of challenges such as illumination, environmental and temporal changes. Examples are in Figure \ref{fig:dataeg}.

\subsubsection{Oxford RobotCar}
The Oxford RobotCar dataset~\cite{RobotCarDatasetIJRR} captures over 100 repetitions of a route through Oxford across a 1 year period. It includes many challenging conditions found in long term localization including illumination, seasonal and temporal changes due to construction. We use a sunny day traverse for the reference database and include common adverse driving conditions using a dusk evening and an overcast day traverse as query datasets. We refer to these as \textit{RobotCar Dusk} and \textit{RobotCar Overcast}.

\subsubsection{Nordland}
Nordland~\cite{Sunderhauf2013} is an established benchmark dataset used for visual place recognition and localization. The dataset includes images from a 728km train journey in all four seasons of the year. The changing conditions and perceptual aliasing throughout makes this a tough and unique VPR dataset. We make use of this dataset by using the summer, spring and fall sets in a combination of query and reference pairs. We refer to these as \textit{Nordland Fall-Spring}, \textit{Nordland Fall-Summer} and \textit{Nordland Spring-Summer}.

\subsubsection{Ford AV}
The Ford AV dataset~\cite{agarwal2020ford} presents a challenging driving dataset largely captured from 2017-2018. It includes images captured from vehicles driven throughout Michigan in a range of times, conditions and environments including freeways, city-centers, university campus and suburban areas. In this work we test on query and corresponding reference datasets from each of these environment types. We refer to these by their 'Log' numbers (and year where applicable), \textit{Ford 1 2017}, \textit{Ford 3}, \textit{Ford 4} and \textit{Ford 1 2022}.

\subsubsection*{Target Localization Performance Values}
In order to train and test our approach on these datasets, we needed to determine what level of VPR performance could reasonably be maintained across each of them. Using HybridNet~\cite{chen2017deep} as our VPR feature extractor, we analysed the recall values from a sweep of sequence lengths, as in Section \ref{subsubsec:trgtperf}, to set the target localization/recall performance for each dataset. These values can be found in Table \ref{tab:datasum}.

\def\scaleone{0.0513}
\def\scaletwo{0.1026}
\def\scalethree{0.0297}
\setlength\tabcolsep{1.5pt}

 \begin{table}
     \centering
     \begin{tabular}{c}  
         \begin{tabular}{ccc}
             \includegraphics[scale=\scaleone]{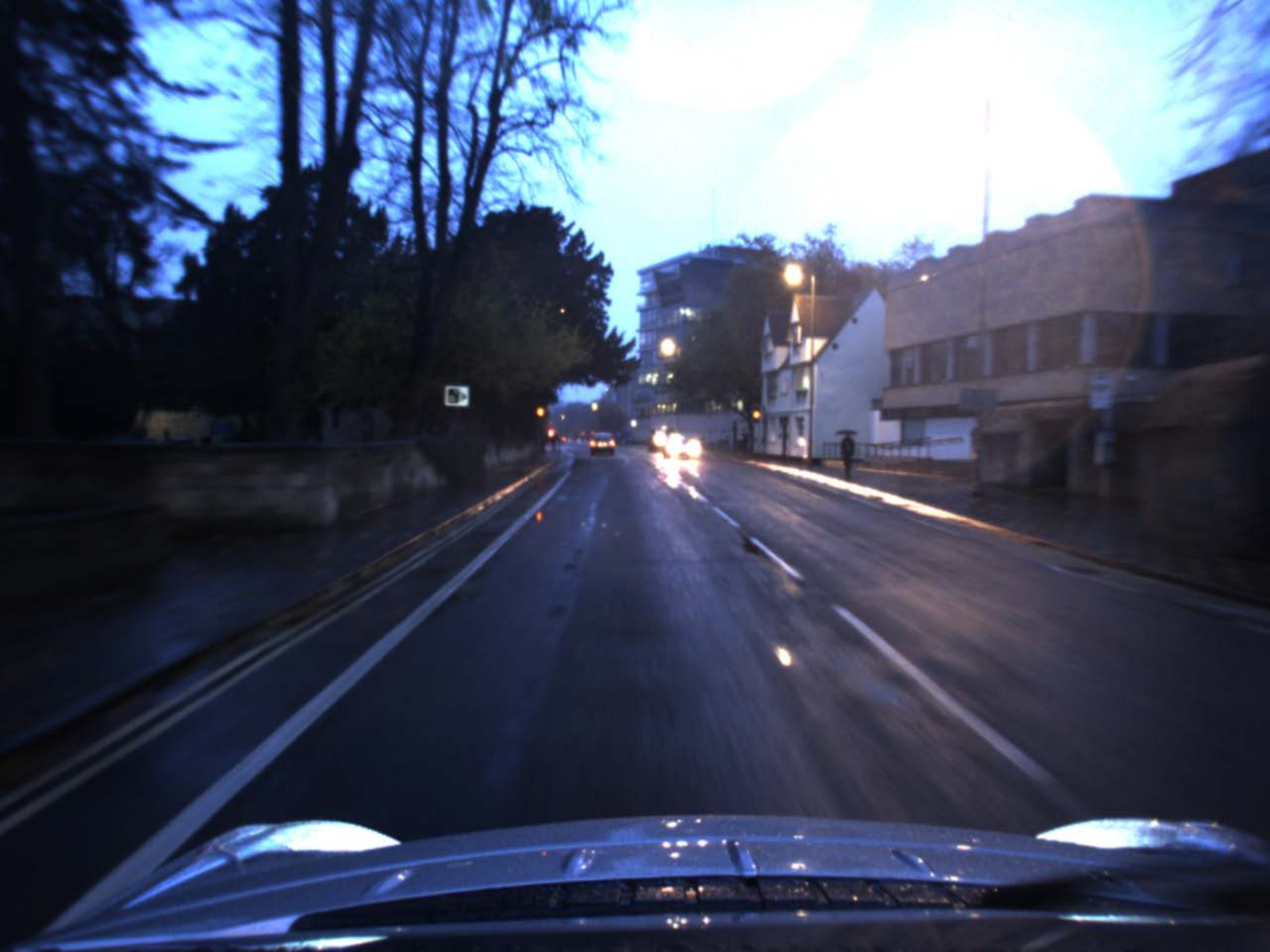} & 
             \includegraphics[scale=\scaleone]{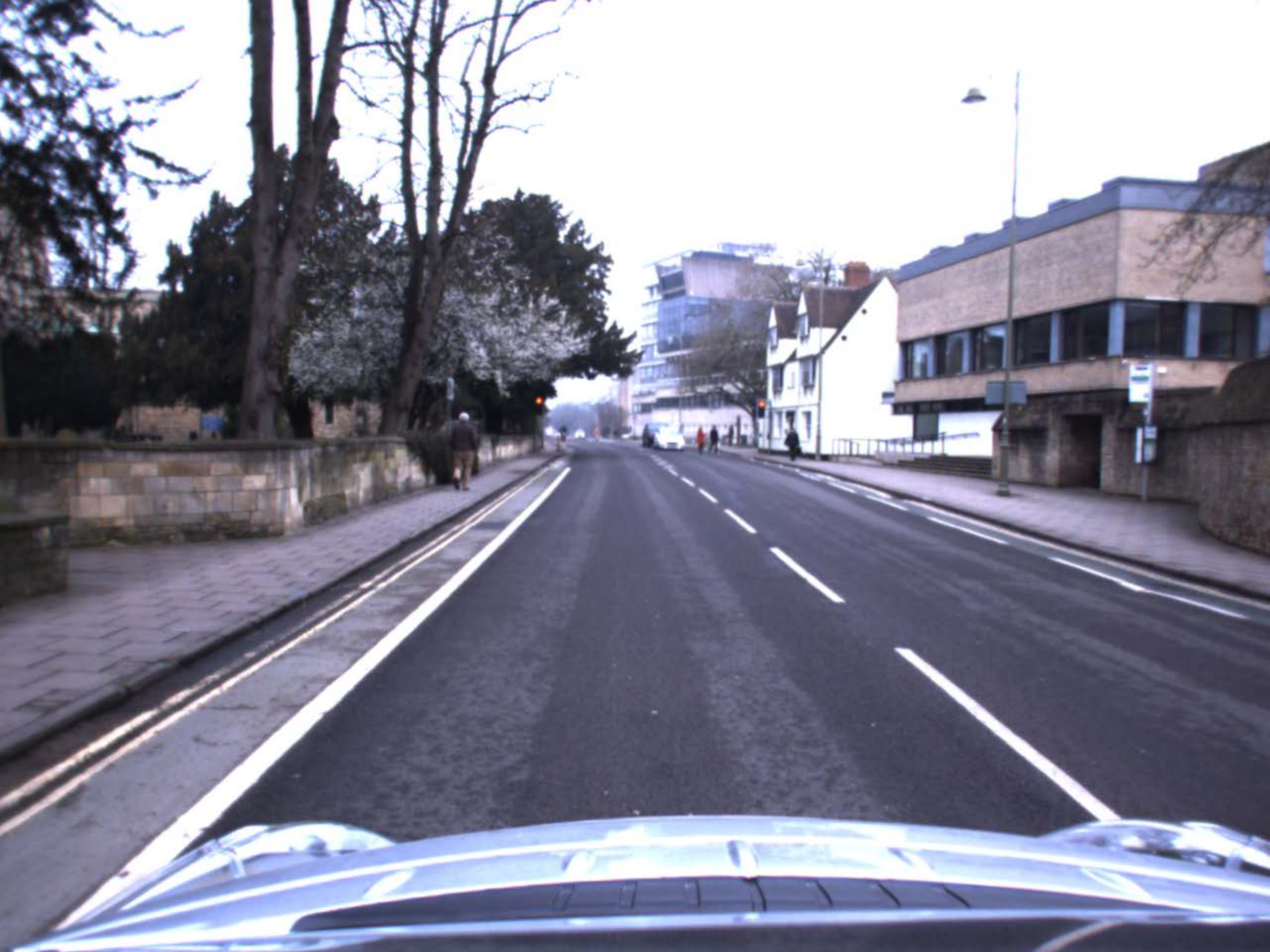} &
             \includegraphics[scale=\scaleone]{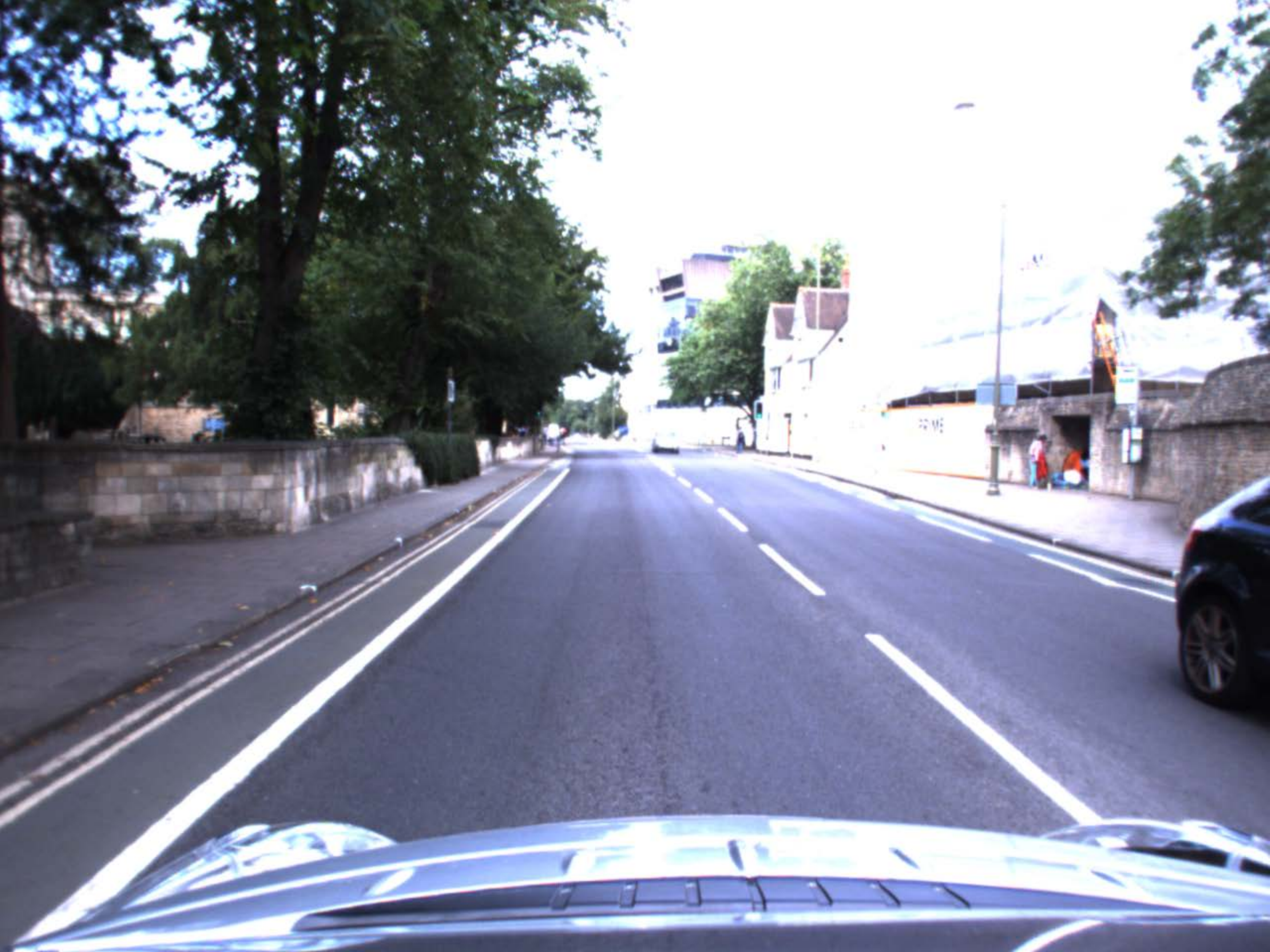} \\
             \includegraphics[scale=\scaletwo]{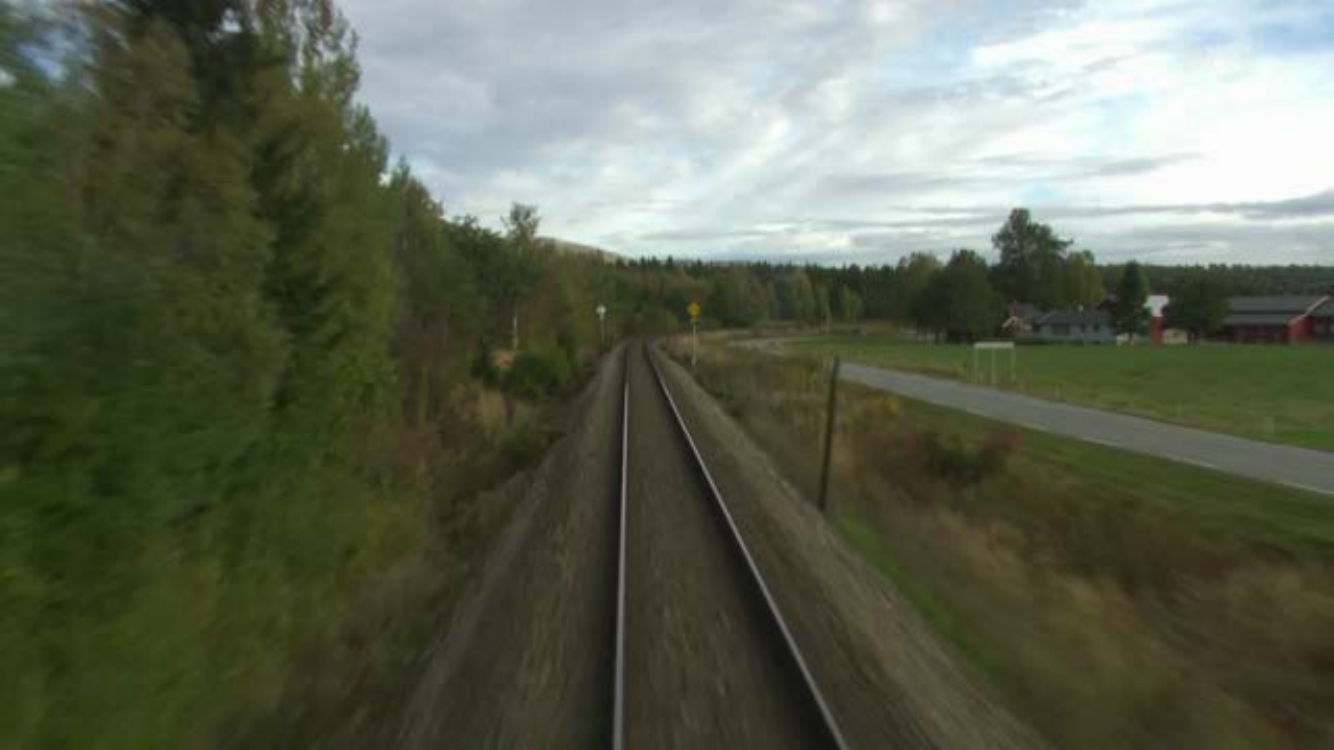} & 
             \includegraphics[scale=\scaletwo]{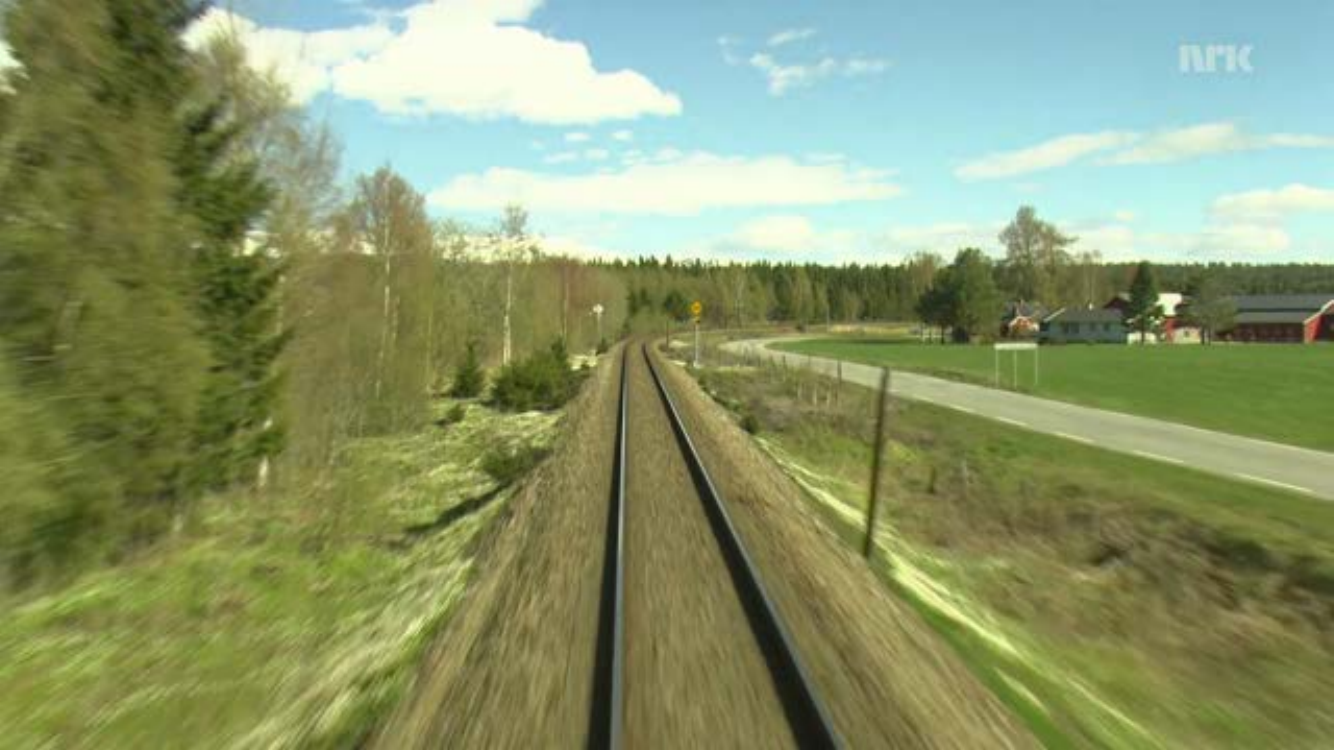} &
             \includegraphics[scale=\scaletwo]{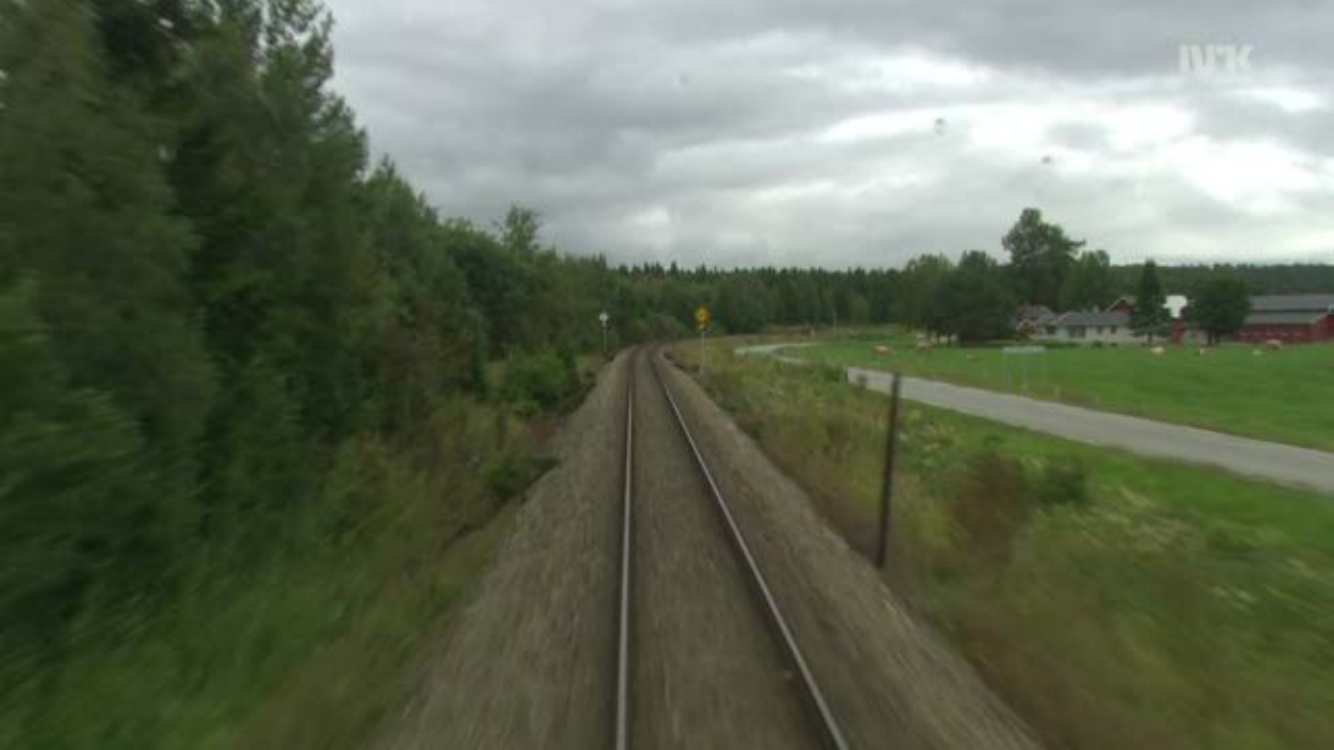} 
         \end{tabular}
    \\  
         \begin{tabular}{cccc}
             \includegraphics[scale=\scalethree]{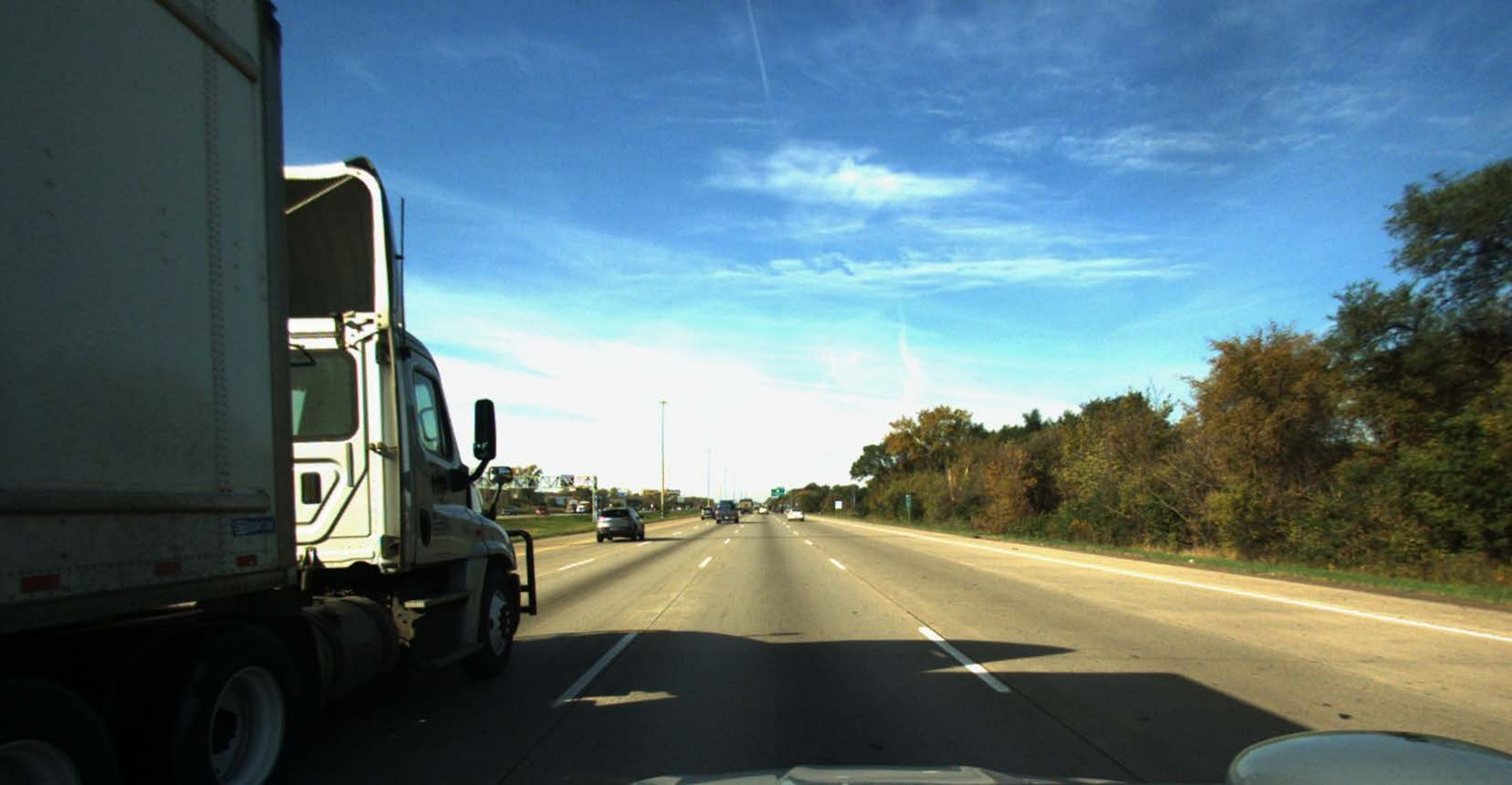} & 
             \includegraphics[scale=\scalethree]{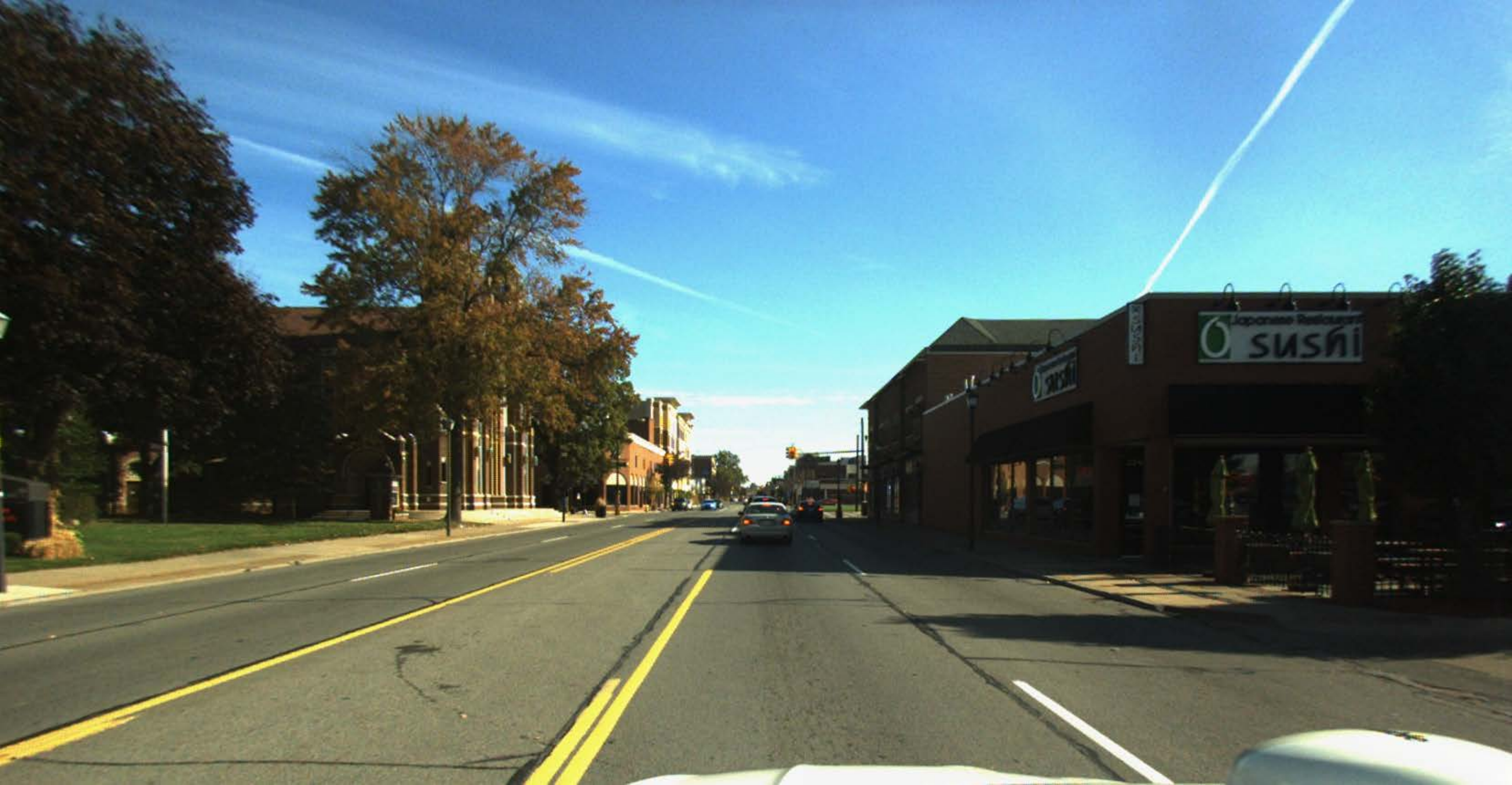}  &
             \includegraphics[scale=\scalethree]{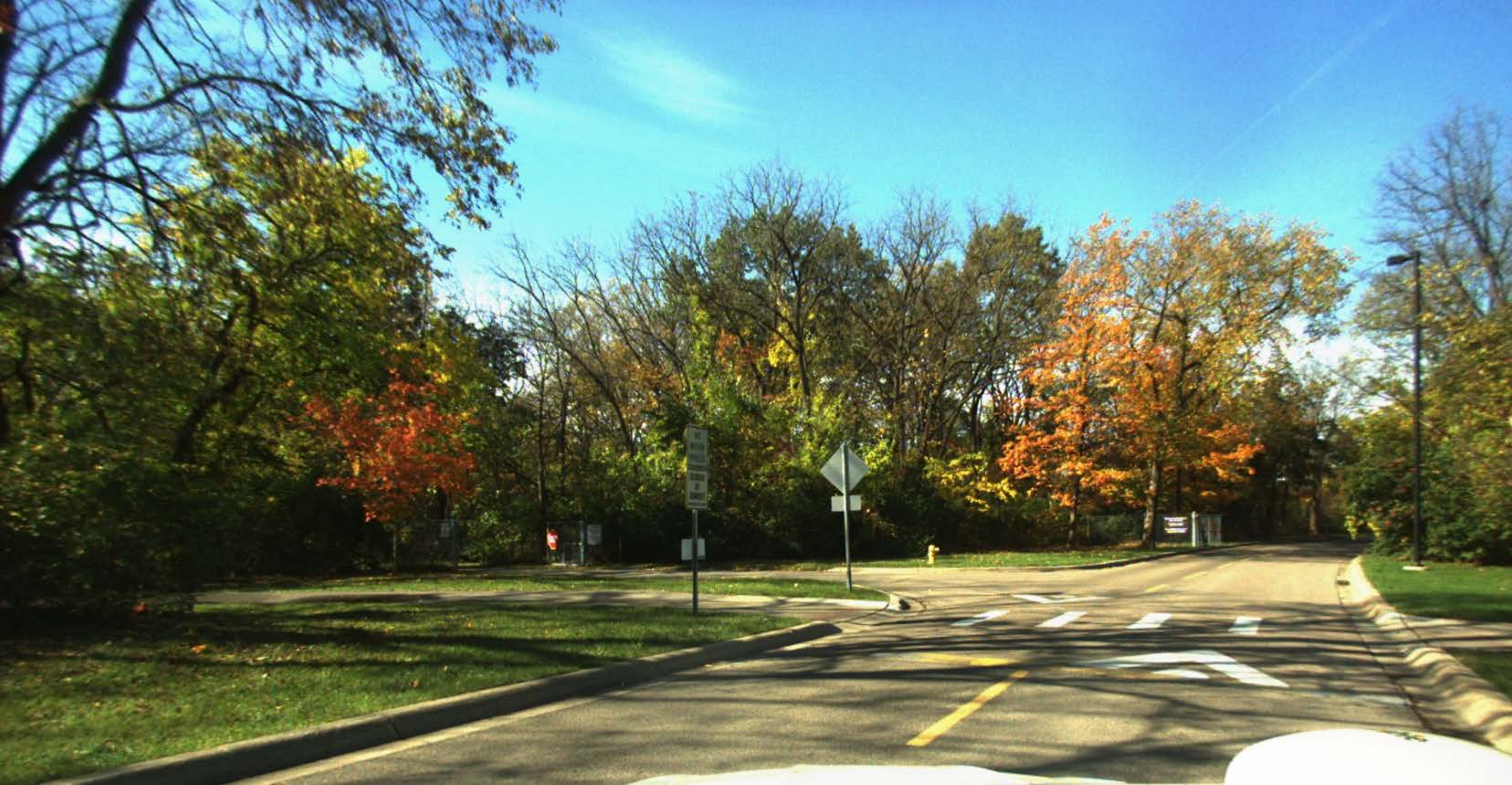} & 
             \includegraphics[scale=\scalethree]{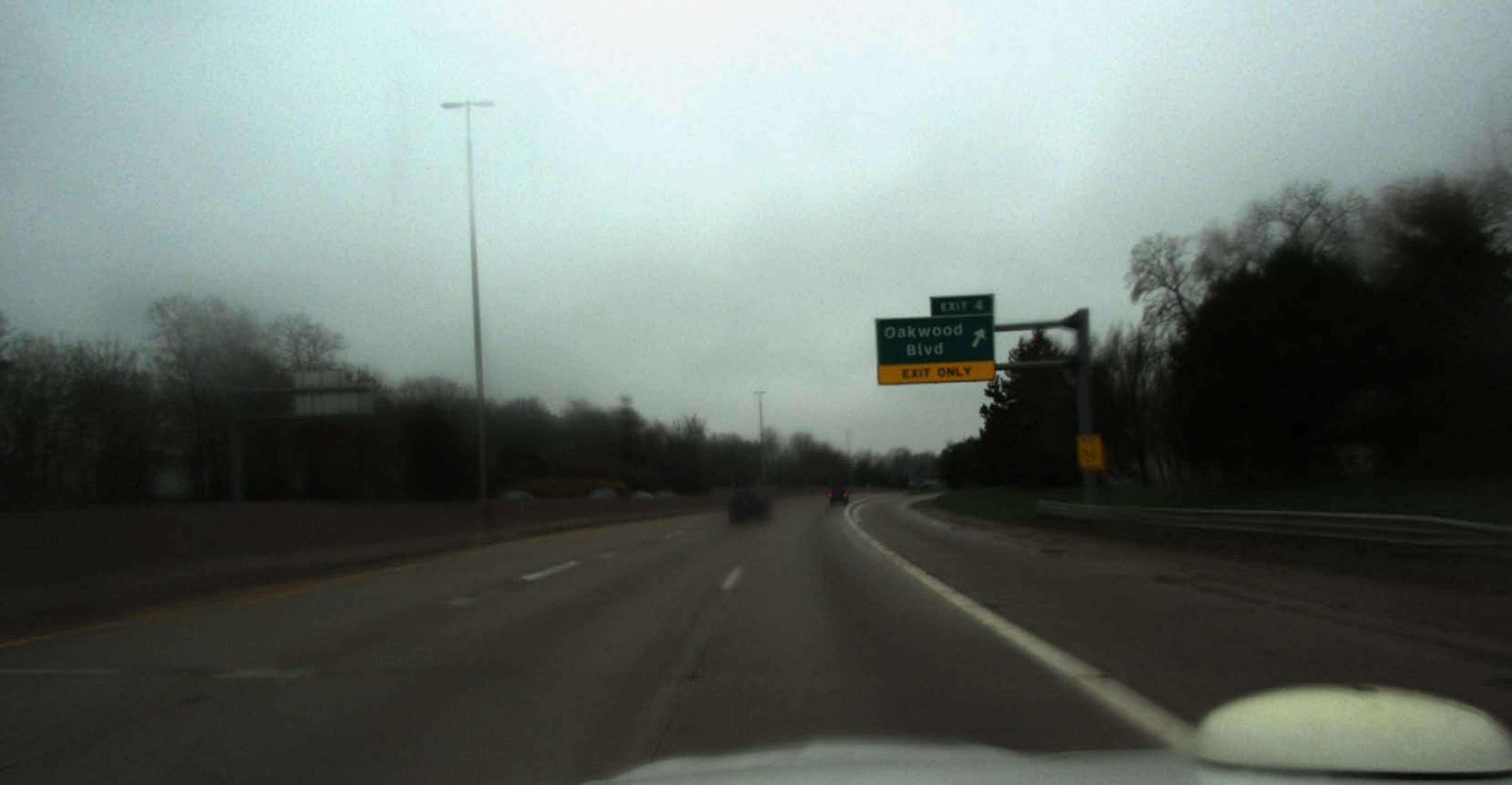} \\
             \includegraphics[scale=\scalethree]{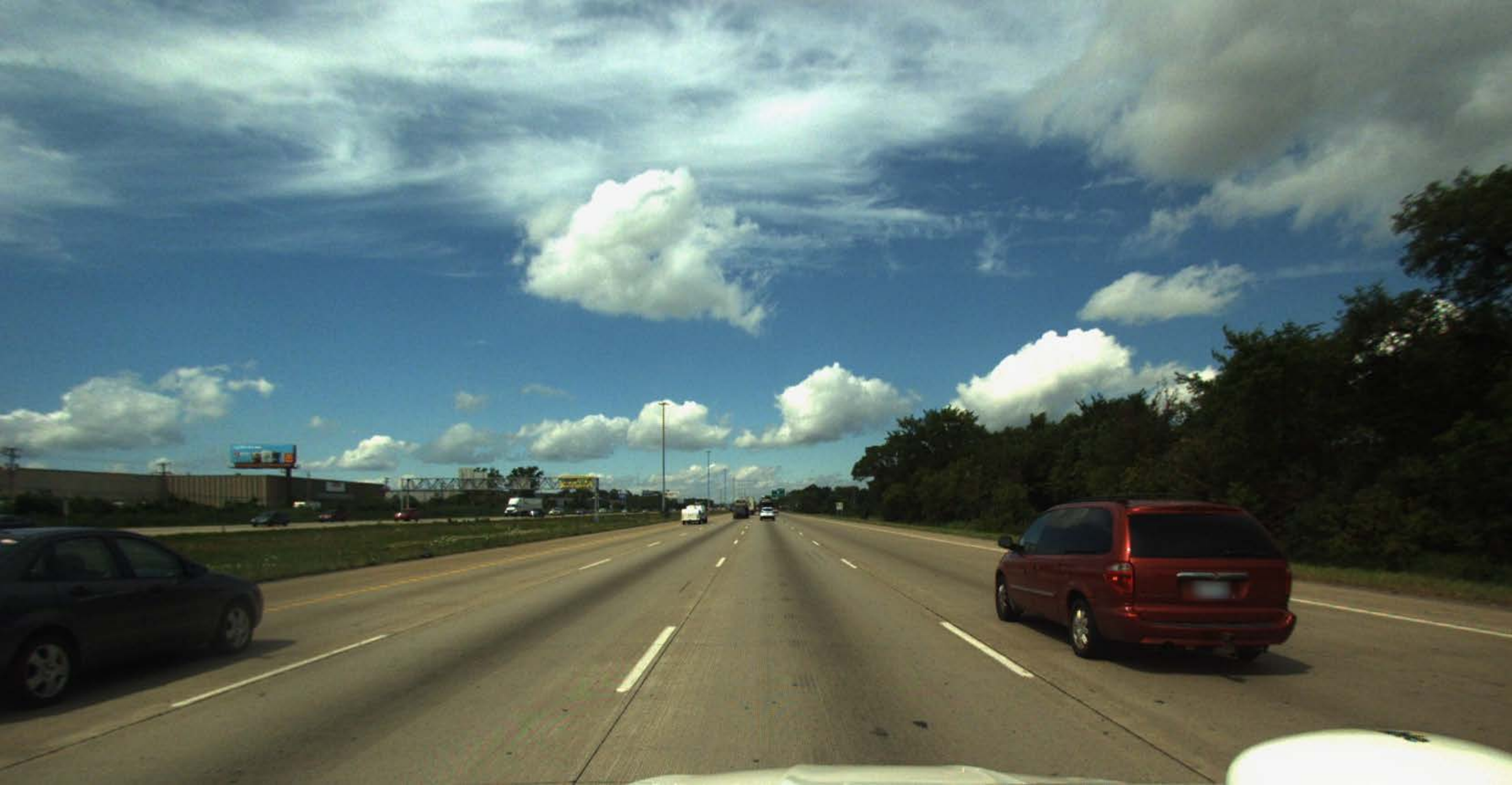}  & 
             \includegraphics[scale=\scalethree]{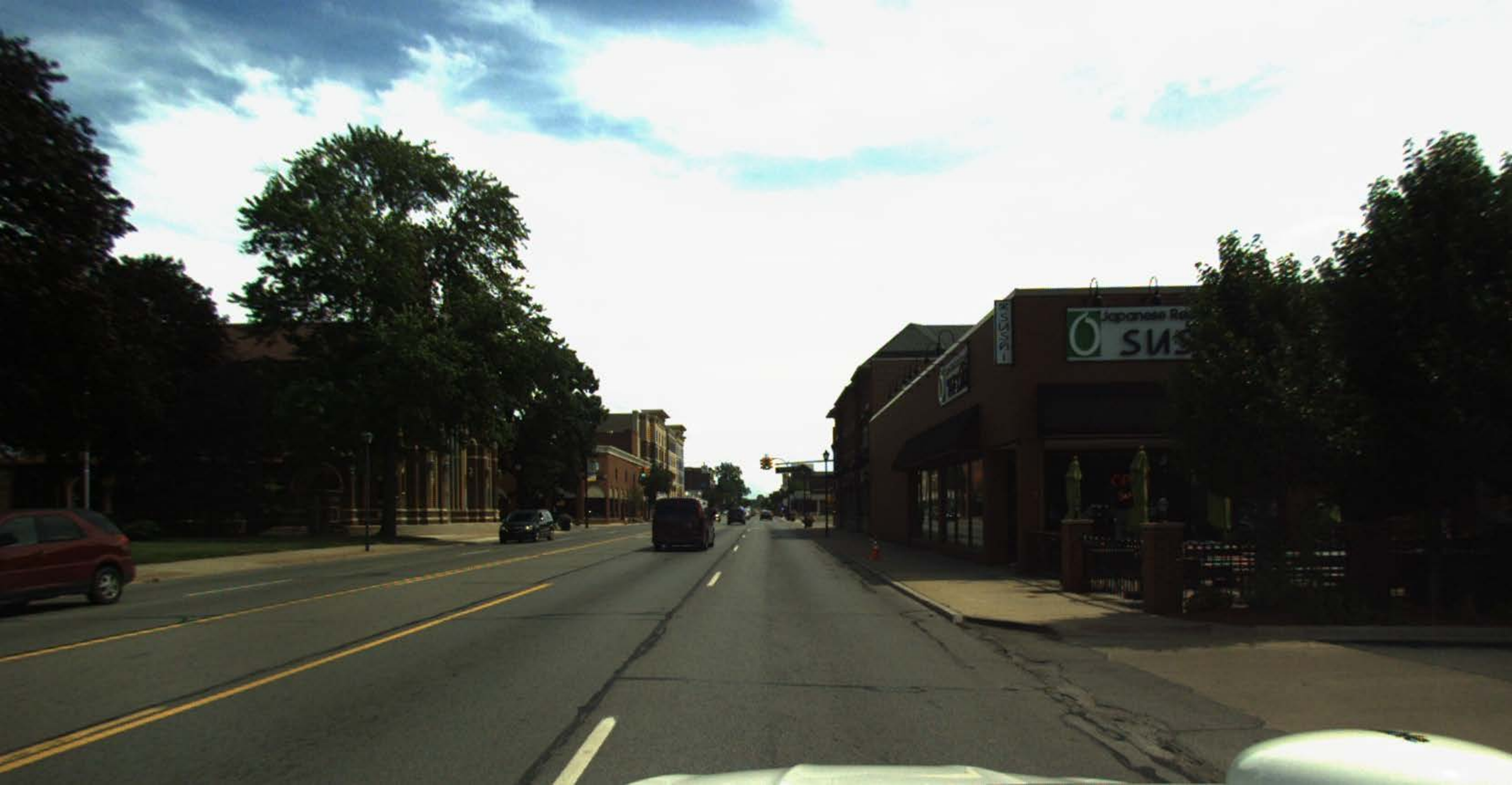}  &
             \includegraphics[scale=\scalethree]{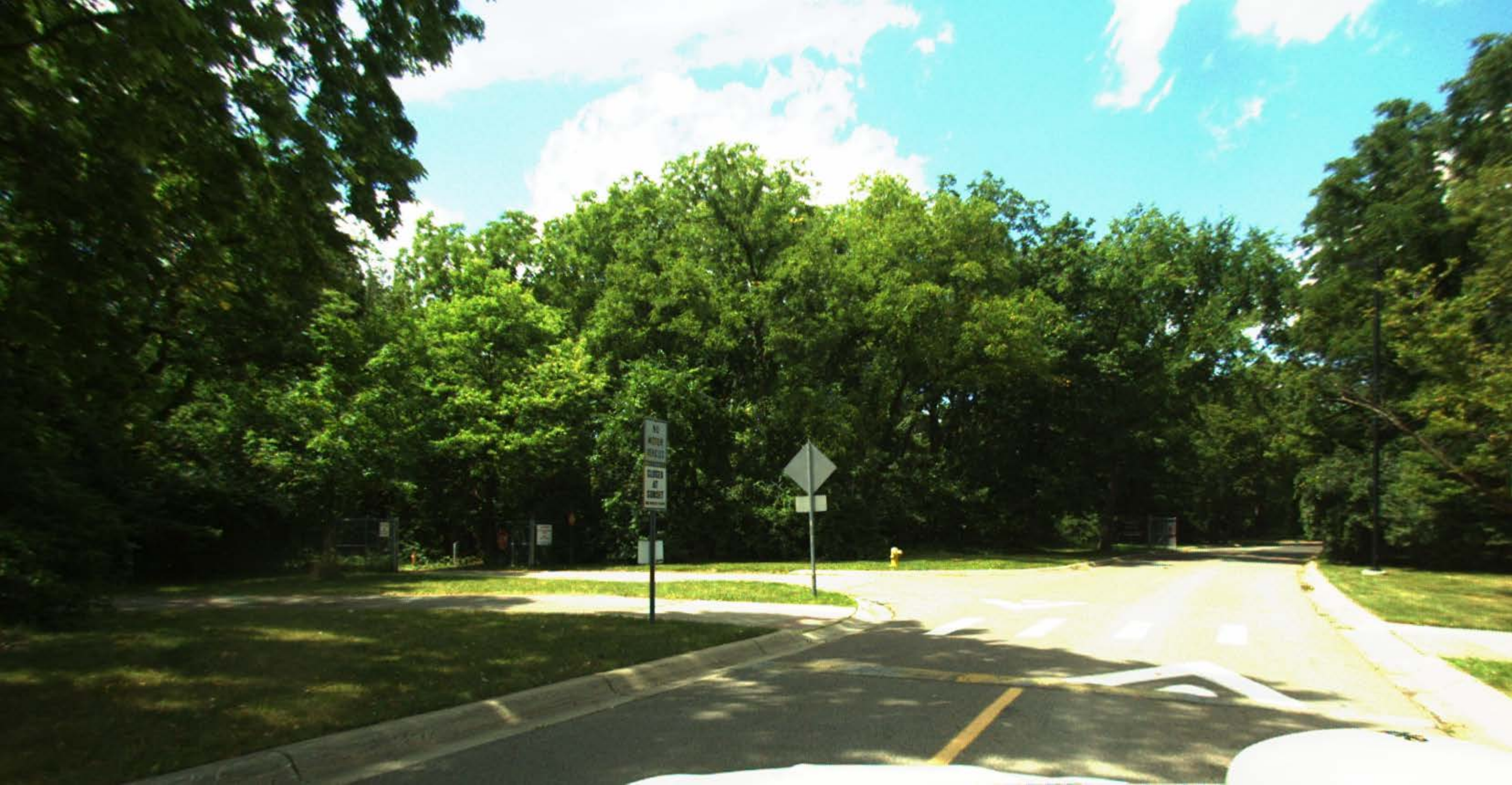} & 
             \includegraphics[scale=\scalethree]{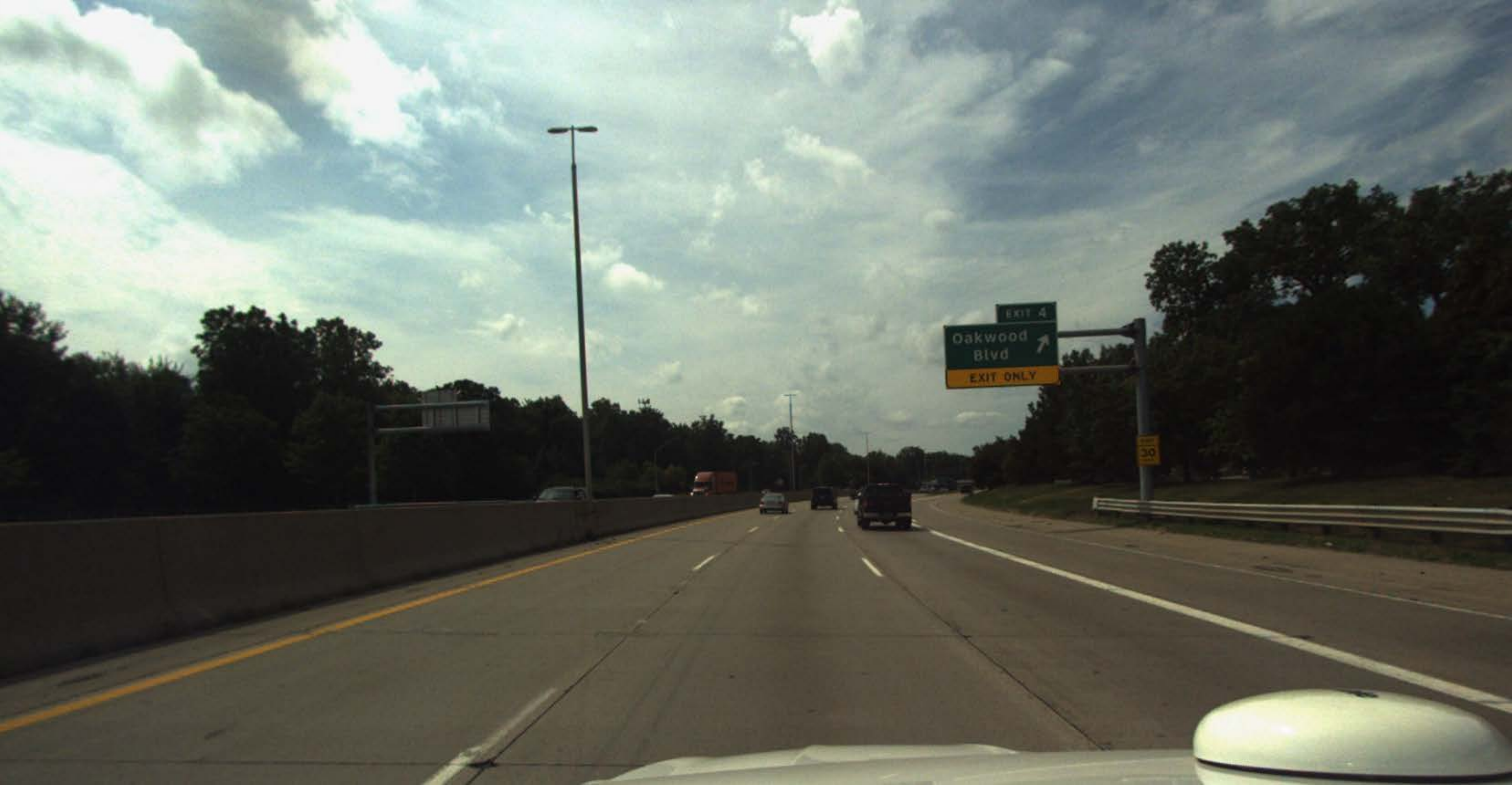} 
         \end{tabular}
     \end{tabular}
     \captionof{figure}{Example images from the datasets used in our work. \textbf{Top Row:} RobotCar Dusk, Overcast and Sunny. \textbf{2nd Row:} Nordland Fall, Spring and Summer. \textbf{3rd and 4th Rows:} query and reference images respectively from the Ford 1 2017, Ford 3, Ford 4 and Ford 1 2022 datasets.}
     \label{fig:dataeg}
     \vspace{-\baselineskip}
 \end{table}

\setlength\tabcolsep{6pt}

\subsection{Experiments}
\label{subsec:paradigm}
To evaluate how effective our proposed approach is at maximizing the proportion of a dataset/environment that exceeds a target recall, we establish the following experiment protocols. In general, for each dataset, we separate images into `chunks' and perform VPR within these `chunks' using different paradigms for selecting a sequence length.

As datasets with uncertain GPS/coarse positions which are also functional as VPR datasets are difficult to find, we manually set the coarse position prior size, $m$, as in Section \ref{subsubsec:trgtperf}. We create `chunks' through the dataset with a step size smaller than $m$ to better represent possible position prior locations and generate more training and test data. All datasets except Nordland have been sampled at roughly 1m intervals. We therefore set the chunk size, $m$, to be 75 images to represent a coarse position estimate with a 75m length for all datasets except in the case for Nordland. We sample these `chunks' from the datasets with a step size of 15 images.

For the purpose of calibrating the proposed MLP regression model, we separate the datasets into training, validation and testing splits of 30\%, 20\% and 50\% respectively. Combined with the selected chunk and step size, the characteristics of the datasets utilised in this work are summarised in Table \ref{tab:datasum}. During training we assess regression performance purely based on the sequence length training signal, $s$, determined using Section \ref{subsubsec:trgtperf}. For localization performance across the test splits we perform VPR for every image within each chunk and calculate the performance for each chunk independently. Testing VPR localization performance in this way could be considered a representation of the chance for localization success at each chunk given that the true position could be any point within the uncertain/coarse position prior.

\setlength\tabcolsep{4.5pt}
\begin{table}
    \centering
    \caption{Summary of Tested Datasets' Characteristics}
    \label{tab:datasum}
    \begin{tabular}{lccccc}
        \toprule
         & & \multicolumn{4}{c}{Chunks}   \\
         \cmidrule(lr{0.75em}){3-6}
        \textbf{Dataset (Target Recall)} & \textbf{Images} & \textbf{Train} & \textbf{Valid} & \textbf{Test} & \textbf{Total} \\
        \midrule
        RobotCar Dusk (50\%) & 3876 & 229 & 153 & 383 & 765 \\
        RobotCar Overcast (50\%) & 3876 & 229 & 153 & 383 & 765 \\
        \midrule
        Ford 1 2017 (75\%) & 5924 & 351 & 234 & 585 & 1170 \\
        Ford 3 (75\%) & 4602 & 274 & 183 & 458 & 915 \\
        Ford 4 (75\%) & 3087 & 184 & 123 & 308 & 615 \\
        Ford 1 2022 (40\%) & 3987 & 238 & 159 & 398 & 795\\
        \midrule
        Nord Fall-Spring (85\%) & 11923 & 711 & 474 & 1185 & 2370 \\
        Nord Fall-Summer (85\%) & 11923 & 711 & 474 & 1185 & 2370 \\
        Nord Spring-Summer (85\%) & 11923 & 711 & 474 & 1185 & 2370 \\
        \bottomrule
    \end{tabular}
    \vspace{-\baselineskip}
\end{table}
\setlength\tabcolsep{6pt}

\begin{table*}
\centering
\begin{minipage}{0.58\linewidth}
\centering
\setlength\tabcolsep{5pt}
    \caption{\underline{\textbf{Sects \%}}: The percentage of `chunks' which meet or exceed the target performance. \underline{\textbf{Len}}: Median sequence length across `chunks'.}
    \begin{tabular}{@{}lcccccc@{}} %
        \toprule
        & \multicolumn{2}{c}{No Sequence} & \multicolumn{2}{c}{Ours}  & \multicolumn{2}{c}{Fixed (Train Len)}\\
        \cmidrule(lr{0.75em}){2-3}
        \cmidrule(lr{0.75em}){4-5}
        \cmidrule(lr{0.75em}){6-7}
        \textbf{Dataset (Target Recall)}  & Sects \% & Len & Sects \% & Len  & Sects \% & Len     \\
        \midrule
        RC Dusk (50\%) & 22.5 & 1 & \textbf{72.3} & 11 & 43.6 & 5 \\
        RC Overcast (50\%)  & 55.6 & 1 & \textbf{77.8} & 9 & 55.6 & 1 \\
        \midrule
        Ford 1 2017 (75\%) & 36.8 & 1 & \textbf{66.7} & 7 & 59.3 & 5 \\
        Ford 3 (75\%)  & 16.2 & 1 & \textbf{51.3} & 13 & 19.0 & 3     \\
        Ford 4 (75\%) & 28.6 & 1 & \textbf{44.2} & 3 & 43.2 & 5        \\
        Ford 1 2022 (40\%) & 24.4 & 1 & \textbf{38.4} & 13  & 29.6 & 3 \\
        \midrule
        Nord Fall Spring (85\%)   & 22.3 & 1 & \textbf{78.8} & 17 & 71.1 & 5 \\
        Nord Fall Summer (85\%) & 52.0 & 1 & \textbf{82.5} & 13 & 76.5 & 5 \\
        Nord Spring Summer (85\%) & 13.0 & 1 & \textbf{81.9} & 15 & 69.3 & 5 \\
        \bottomrule
    \end{tabular}
    \label{tab:main}
\setlength\tabcolsep{6pt}
\end{minipage}
\begin{minipage}{0.01\linewidth}
\begin{tabular}{c}

\end{tabular}
    
\end{minipage}
\begin{minipage}{0.39\linewidth}
\centering
        \begin{tabular}{c}
           \includegraphics[width=0.7\linewidth]{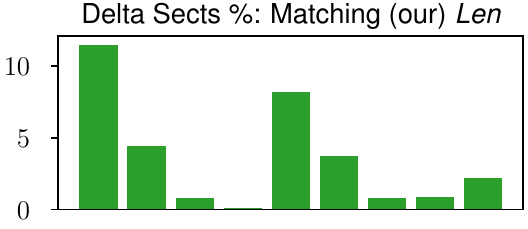}  \\
           \hdashline
           \includegraphics[width=0.76\linewidth]{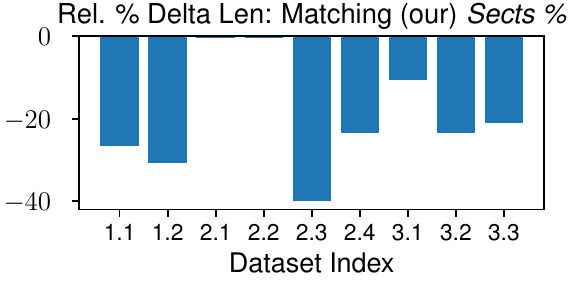} 
        \end{tabular}
        \vspace{-\baselineskip}
        \captionof{figure}{Delta improvements of our approach over fixed sequence lengths when matching its median sequence length (\textbf{Top}) \underline{or} consistency (\textbf{Bottom}) metrics. Dataset indices follow order in Table \ref{tab:main}.}
        \label{fig:oracomps}
\end{minipage}
\vspace{-2.5\baselineskip}
\end{table*}

\subsection{Metrics}
\label{subsec:metrics}
As mentioned, throughout this work we use the recall@1 metric widely adopted for assessing the localization performance of place recognition systems. To address the ultimate objective for this work of maintaining consistent VPR performance throughout an environment, we simply use the percentage of dataset `chunks' that meet or exceed the target recall@1. Further to this point, we use the median sequence length across all `chunks' to quantify how efficiently the system is able to achieve this localization consistency.

\subsection{Comparison Methods}
\label{subsec:baselines}
For evaluation we compare our approach to four alternatives which demonstrate the possible performance from using a fixed sequence length throughout the entire dataset. To provide fair comparisons we perform VPR within the reduced search space from the coarse/uncertain position prior for \textit{all} methods. The first comparison uses single image VPR (no sequences) and establishes the baseline for performance (Table \ref{tab:main}, col. 1). Secondly we compare to the performance of using the smallest fixed sequence length from the training set which on average achieves the target localization performance (Table \ref{tab:main}, col. 3). Finally we compare to fixed sequence length approaches that either match the consistency metric (\% of `chunks'; Fig. \ref{fig:oracomps}, bottom) \underline{or} median sequence length (Fig. \ref{fig:oracomps}, top) of our dynamic sequence length approach.

\subsection{Network Parameters and Optimisation}
We optimised hyperparameters to the following values. We set the coarse position prior, $m = 75$. HybridNet feature vectors have dimensions, $n = 4096$. We set hidden layers $L = 3$, hidden layer neurons $N = 128$, AMI threshold 
$\alpha= 0.125$, loss values $\beta = 1$ and $\gamma = 0.01$, and the number of input $p$ is unique to each dataset depending on the number of features above the AMI threshold in the training sets.

\begin{figure}
    \centering
    \includegraphics[width=0.85\linewidth]{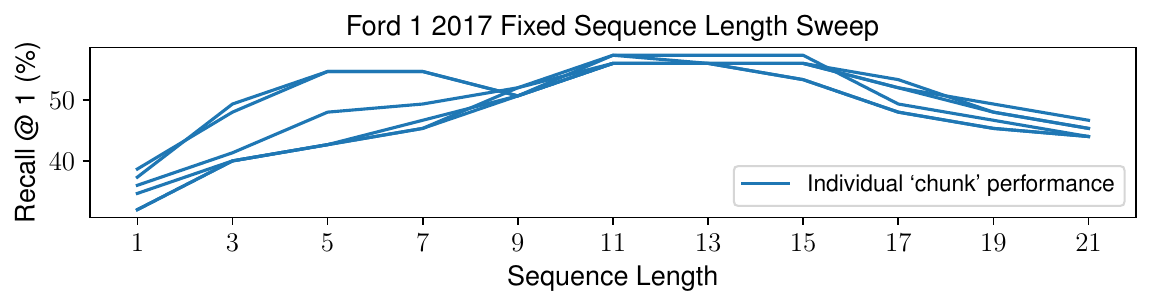}
    \caption{`Chunks' from Ford 1 2017 showing longer sequence lengths can sometimes degrade performance.}
    \label{fig:longseqlens}
    \vspace{-\baselineskip}
\end{figure}

\section{Results}
\label{RESULTS}

In this section, we present results that demonstrate the advantage of using our sequence length modulation approach for consistently exceeding a target localization performance. Table \ref{tab:main} and Fig. \ref{fig:oracomps}-\ref{fig:RCD} provide quantitative evidence to support the benefit of our approach towards this objective, whilst Tables \ref{tab:ftsel}, \ref{tab:gen} and \ref{tab:ftex} provide further analysis of its properties.

\subsection{Exceeding Target Localization Performance}
\label{subsec:resultsmain}
We start by analysing the results in Table \ref{tab:main} and Fig.~\ref{fig:oracomps} comparing our approach to the alternative sequence length strategies described in Section \ref{subsec:baselines}. Importantly, it can be seen in the table that using any sequence-based VPR method typically results in higher percentage of `chunks' exceeding the target localization performance over single image VPR (col. 1). Given all VPR was performed using the coarse position priors, this demonstrates that the majority of improvement from our approach is coming from modulating sequence length rather than the position prior itself. Fig. \ref{fig:front} illustrates how much our approach can modulate sequence length over a dataset, using RobotCar Dusk, while  Fig. ~\ref{fig:longseqlens} highlights the scenario where simply setting long fixed sequence lengths can sometimes degrade performance.

First we compare to the smallest fixed sequence length that exceeds the target localization performance across the training split used to fit our MLP regression model (col. 3). Across all datasets, our approach (col. 2) is able to exceed the target performance for a higher percentage of `chunks' than this comparison. This result clearly demonstrates the challenges this work attempts to address; that traditional localization metrics do not guarantee consistent performance throughout a dataset; and that fixed sequence lengths are not guaranteed to be suitable/adequate across entire datasets. Our approach is able to make more effective use of the training data by dynamically modulating the sequence length to exceed performance throughout a greater proportion of the dataset. Notably the median sequence length our approach uses is generally higher, likely to accommodate sections of the dataset where a longer sequence length is necessary.

\begin{figure}
    \centering
    \includegraphics[width=\linewidth]{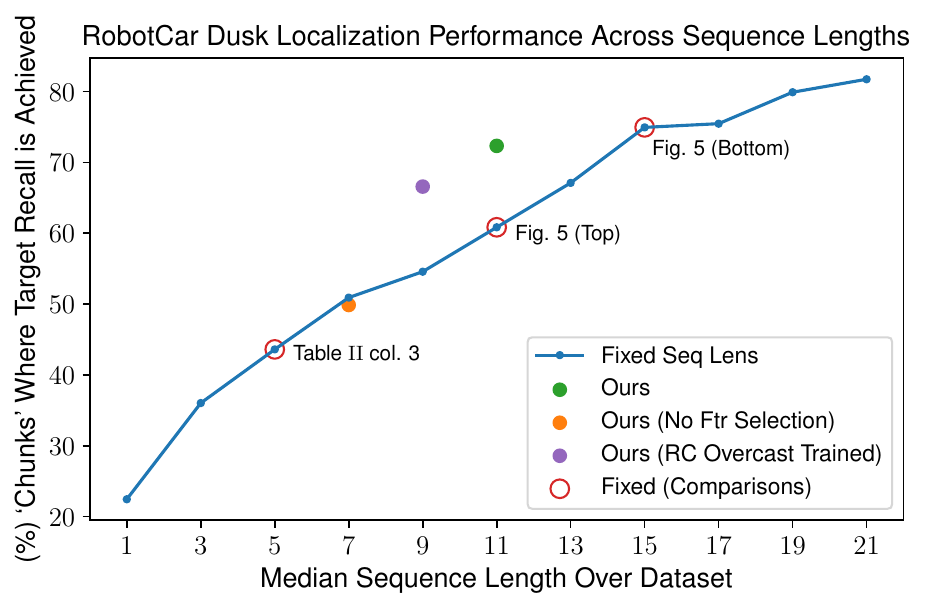}
    \caption{Localization performance of varying sequence lengths and ablations of our approach for the RobotCar Dusk dataset.}
    \label{fig:RCD}
    \vspace{-1\baselineskip}
\end{figure}

Next we compare to the fixed sequence lengths that either match the consistency or median sequence length metrics of our approach (Fig. \ref{fig:oracomps}). These comparisons are necessary for the performance analysis of our approach, however it is important to note that they are not practical methods for choosing fixed sequence lengths as they use `oracle' type information not available at test time. When comparing to using the fixed sequence length which most closely matches the percentage of `chunks' (\textit{Sects \%}) from our approach, Fig. \ref{fig:oracomps} shows our approach consistently achieves a lower median sequence length. This demonstrates that our approach is able to achieve this localization consistency with lower latency and therefore more efficiency. Furthermore, using the fixed sequence length matching the median length of our approach (\textit{Len}) clearly shows that our approach exceeds the target performance for a higher percentage of dataset `chunks'.

We provide Fig. \ref{fig:RCD} as a visualisation of how the performance of our approach compares to the entire sweep of fixed sequence lengths for the RobotCar Dusk dataset. This figure can be interpreted as an operating curve for the sweep of sequence lengths. Therefore to prove that our approach is useful towards the objective of this work and more effective than fixed sequence lengths, we show that the operating point of our approach lies above this operating curve.

\setlength\tabcolsep{3pt}

\begin{table}
\centering
\caption{Removing Feature Selection}
\begin{tabular}{@{}lcccccc} %
\toprule
& \multicolumn{2}{c}{Ours (\textbf{no feat. sel})} & \multicolumn{2}{c}{Ours} & \multicolumn{2}{c}{Fixed (Match Len)} \\ 

\cmidrule(lr{0.75em}){2-3}
\cmidrule(lr{0.75em}){4-5}
\cmidrule(lr{0.75em}){6-7}
\textbf{Dataset}  & Sects \% & Len & Sects \% & Len & Sects \% & Len \\
\midrule
RC Dusk & 49.9 & 7 & \textbf{72.3} & 11 & 50.9 & 7 \\
RC Overcast & 69.2 & 5 & \textbf{77.8} & 9 & 68.1 & 5 \\
\midrule
Ford 1 2017 & 71.3 & 13 & 66.7 & 7 & \textbf{78.1} & 13 \\ 
Ford 3 & 22.0 & 5 & \textbf{51.3} & 13 & 25.3 & 5 \\ 
Ford 4 & 40.3 & 3 & \textbf{44.2} & 3 & 36.0 & 3 \\
Ford 1 2022 & 32.2 & 7 & \textbf{38.4} & 13 & 30.7 & 7 \\
\midrule
Nord Fall Spr. & 76.2 & 15 & \textbf{78.8} & 17 & 76.5 & 15 \\
Nord Fall Sum. & \textbf{82.6} & 13 & 82.5 & 13 & 81.6 & 13 \\
Nord Spr. Sum. & 79.0 & 13 & \textbf{81.9} & 15 & 79.3 & 13\\
\bottomrule
\end{tabular}
\label{tab:ftsel}
\vspace{-\baselineskip}
\end{table}

\setlength\tabcolsep{6pt}

\subsection{Contribution of Correlated Feature Selection}
In our approach we provide a method for curating the most correlated features to use as the MLP regression model input (Section \ref{subsec:optfeats}). Fig. \ref{fig:corrLeRelu}a shows that there are particular features which are harmful to sequence length regression according to AMI scores. However, we also provide Table \ref{tab:ftsel} and Fig. \ref{fig:RCD} to demonstrate this experimentally. Table \ref{tab:ftsel} clearly shows that removing optimal feature selection from our approach typically lowers the percentage of dataset `chunks' that exceed the target localization performance. Furthermore, using the fixed comparison which matches the median sequence length shows there would be minimal or no improvement over using a fixed sequence length. This importance of feature selection is reiterated in Fig. \ref{fig:RCD} where the performance of our system changes from above to below the operating curve when removing this process.

\begin{table}
\centering
\caption{Generalization to Separate Datasets}
\begin{tabular}{@{}lcccccc} %
\toprule
\textbf{Test Data $\rightarrow$} & \multicolumn{2}{c}{Ford 1 2017} & \multicolumn{2}{c}{Ford 3}  & \multicolumn{2}{c}{Ford 4}  \\ %

\cmidrule(lr{0.75em}){2-3}
\cmidrule(lr{0.75em}){4-5}
\cmidrule(lr{0.75em}){6-7}
\textbf{Train Data $\downarrow$}  & Sects \% & Len & Sects \% & Len  & Sects \% & Len  \\ %
\midrule
Ford 1 2017 & 66.7 & 7 & 43.0 & 9 & 62.3 & 15 \\ %
Ford 3 & 80 & 17 & 51.3 & 13 & 68.2 & 15 \\ %
Ford 4 & 57.6 & 7 & 50 & 21 & 44.2 & 3 \\ %
All & 80.3 & 19 & 53.3 & 17 & 65.3 & 17 \\
\bottomrule
\end{tabular}
\label{tab:gen}
\vspace{-1.5\baselineskip}
\end{table}

\subsection{Generalization to Other Datasets}
For most experiments we trained an MLP regression model for each dataset using a small calibration set and tested on a separate unseen section from the same dataset. Table \ref{tab:gen} provides results which removes this limitation and explores the generalization capabilities of our approach. It shows the performance of each Ford dataset trained model on the other Ford datasets. We include results for the Ford 1 2017, Ford 3 and Ford 4 datasets which share the same target localization performance and train a single model combining all calibration sets.

Interestingly, Table \ref{tab:gen} shows that when our approach is tested on completely different datasets to its training, it tends to select higher sequence lengths overall. This appears to generally result in a higher percentage of dataset `chunks' that exceed the target localization performance. However, it is likely that the sequence length selections are not as appropriate as on the trained dataset. This is evidenced in column 1 where the model trained on Ford 4 has the same median sequence length but a lower percentage of `chunks' than the model trained specifically for Ford 1 2017.

In addition, we include a data point in Fig. \ref{fig:RCD} which demonstrates the generalizability of our approach when train and test queries are captured under different conditions. In the figure we show the performance of our approach trained for the RobotCar Overcast dataset when tested on RobotCar Dusk. This model still performs above the operating curve and therefore is beneficial over a fixed sequence length, despite exceeding the target performance in a lower percentage of `chunks' than one specifically trained for RobotCar Dusk.

\setlength\tabcolsep{3pt}
\begin{table}
\centering
\caption{Using SOTA VPR Feature Extractors}
\begin{tabular}{@{}lcccccc} %
\toprule
& & & \multicolumn{2}{c}{NetVLAD~\cite{arandjelovic2016netvlad}} & & \\
\cmidrule(lr{0.75em}){2-7}
& \multicolumn{2}{c}{No Seq} & \multicolumn{2}{c}{Ours} & \multicolumn{2}{c}{Fixed (Match Len)}\\ 

\cmidrule(lr{0.75em}){2-3}
\cmidrule(lr{0.75em}){4-5}
\cmidrule(lr{0.75em}){6-7}
\textbf{Dataset (Target)}  & Sects \% & Len & Sects \% & Len & Sects \% & Len \\
\midrule
Ford 1 2017 (75\%) & 52.1 & 1 & 69.6 & 7 & \textbf{73.8} & 7 \\ 
Ford 3 (75\%) & 30.6 & 1 & \textbf{54.1} & 7 & 53.3 & 7 \\ 
Ford 4 (75\%) & 35.0 & 1 & 39.0 & 3 & \textbf{43.2} & 3 \\
Ford 1 2022 (40\%) & 23.1 & 1 & 39.4 & 13 & \textbf{42.5} & 13 \\ 
\midrule
& & & \multicolumn{2}{c}{SALAD~\cite{izquierdo2023optimal}} & &\\
\cmidrule(lr{0.75em}){2-7}
& \multicolumn{2}{c}{No Seq} & \multicolumn{2}{c}{Ours} & \multicolumn{2}{c}{Fixed (Match Len)}\\ 

\cmidrule(lr{0.75em}){2-3}
\cmidrule(lr{0.75em}){4-5}
\cmidrule(lr{0.75em}){6-7}
\textbf{Dataset (Target)}  & Sects \% & Len & Sects \% & Len & Sects \% & Len \\
\midrule
Ford 1 2017 (85\%) & 65.8 & 1 & \textbf{87.0} & 3 & 77.3 & 3 \\ 
Ford 3 (85\%) & 34.3 & 1 & 42.4 & 5 & \textbf{45.6} & 5 \\ 
Ford 4 (85\%) & 41.9 & 1 & 43.8 & 3 & \textbf{47.7} & 3 \\
Ford 1 2022 (50\%) & 36.4 & 1 & 48.0 & 5 & \textbf{50.5} & 5 \\ 
\bottomrule
\end{tabular}
\label{tab:ftex}
\vspace{-2\baselineskip}
\end{table}

\setlength\tabcolsep{6pt}

\subsection{Performance With SOTA Feature Extractors}
As described earlier, our proposed approach is particularly suited to VPR feature extractors that maintain some spatial relationship within their features across consecutive images. For completeness and to demonstrate this point we include results in Table \ref{tab:ftex} using NetVLAD~\cite{arandjelovic2016netvlad} and SALAD~\cite{izquierdo2023optimal} as the VPR feature extractors across the Ford datasets. Table \ref{tab:ftex} clearly shows that using our approach with these feature extractors typically results in a lower percentage of `chunks' exceeding the target localization performance compared to the equivalent fixed sequence length. To further support that this is a result of the particular features, it was observed that the typical AMI score for feature correlation was two orders of magnitude lower than the AMI scores observed for the HybridNet features ($\approx0.001$ vs $\approx0.1$). This is likely because both NetVLAD and SALAD cluster features and therefore remove spatial relationships. Using the fine-tuned DINOv2 transformer without the clustering from SALAD may present a feature extractor more compatible with our approach.

\section{Conclusion}
\label{DISCUSSIONANDFUTURE WORK}

In this work we have addressed the challenge of exceeding a target localization performance by modulating VPR sequence length using an MLP regression model. The results have demonstrated that the approach is able to dynamically select sequence lengths to exceed a target performance level over a greater proportion of each dataset than fixed sequence lengths. Thereby mitigating some of the latency and computation costs, and potential performance degradation from simply setting overly long fixed sequence lengths. In addition, they demonstrated the effectiveness of curating features from pretrained networks, the utility of non-SOTA feature extractors with nuanced properties, and our approach offers some generalizability to unseen query conditions.

This work aims to emphasise and support the importance of more practical performance metrics in the field of robotics. Showing that, while important, consistent and reliable performance is not always represented by typical metrics. Using visual place recognition, the proposed approach was shown to more effectively maintain localization performance which is critical when other positioning sensors become unreliable.

\maxdeadcycles=1000

\section*{Acknowledgement}
The authors acknowledge support and funding from the Ford Motor Company through the Ford-QUT Alliance as well as continued support from the Queensland University of Technology (QUT) through the Centre for Robotics.

\bibliographystyle{IEEEtran}
\bibliography{References}

\begin{thebibliography}{10}
\providecommand{\url}[1]{#1}
\csname url@rmstyle\endcsname
\providecommand{\newblock}{\relax}
\providecommand{\bibinfo}[2]{#2}
\providecommand\BIBentrySTDinterwordspacing{\spaceskip=0pt\relax}
\providecommand\BIBentryALTinterwordstretchfactor{4}
\providecommand\BIBentryALTinterwordspacing{\spaceskip=\fontdimen2\font plus
\BIBentryALTinterwordstretchfactor\fontdimen3\font minus
  \fontdimen4\font\relax}
\providecommand\BIBforeignlanguage[2]{{%
\expandafter\ifx\csname l@#1\endcsname\relax
\typeout{** WARNING: IEEEtran.bst: No hyphenation pattern has been}%
\typeout{** loaded for the language `#1'. Using the pattern for}%
\typeout{** the default language instead.}%
\else
\language=\csname l@#1\endcsname
\fi
#2}}

\bibitem{VPR2023Tutorial}
S.~Schubert, P.~Neubert, S.~Garg, M.~Milford, and T.~Fischer, ``Visual place
  recognition: A tutorial,'' \emph{IEEE Robotics \& Automation Magazine}, pp.
  2--16, 2023.

\bibitem{barros2021place}
T.~Barros, R.~Pereira, L.~Garrote, C.~Premebida, and U.~J. Nunes, ``Place
  recognition survey: An update on deep learning approaches,'' \emph{arXiv
  preprint arXiv:2106.10458}, 2021.

\bibitem{lowry2015visual}
S.~Lowry~\textit{et al.}, ``Visual place recognition: A survey,'' \emph{ieee
  transactions on robotics}, vol.~32, no.~1, pp. 1--19, 2015.

\bibitem{schubert2021makes}
S.~Schubert and P.~Neubert, ``What makes visual place recognition easy or
  hard?'' \emph{arXiv preprint arXiv:2106.12671}, 2021.

\bibitem{hausler2022improving}
S.~Hausler~\textit{et al.}, ``Improving worst case visual localization coverage
  via place-specific sub-selection in multi-camera systems,'' \emph{IEEE
  Robotics and Automation Letters}, vol.~7, no.~4, 2022.

\bibitem{Carson2022Integrity}
H.~Carson, J.~J. Ford, and M.~Milford, ``Predicting to improve: Integrity
  measures for assessing visual localization performance,'' \emph{IEEE Robotics
  and Automation Letters}, vol.~7, no.~4, pp. 9627--9634, 2022.

\bibitem{saito2013mobile}
T.~Saito and Y.~Kuroda, ``Mobile robot localization using multiple observations
  based on place recognition and gps,'' in \emph{IEEE International Conference
  on Robotics and Automation}, 2013, pp. 1548--1553.

\bibitem{vysotska2015efficient}
O.~Vysotska, T.~Naseer, L.~Spinello, W.~Burgard, and C.~Stachniss, ``Efficient
  and effective matching of image sequences under substantial appearance
  changes exploiting gps priors,'' in \emph{IEEE international conference on
  robotics and automation (ICRA)}, 2015, pp. 2774--2779.

\bibitem{tomiṭua2022sequence}
M.-A. Tomiṭ{\u{a}}, M.~Zaffar, B.~Ferrarini, M.~J. Milford, K.~D.
  McDonald-Maier, and S.~Ehsan, ``Sequence-based filtering for visual
  route-based navigation: Analyzing the benefits, trade-offs and design
  choices,'' \emph{IEEE Access}, vol.~10, pp. 81\,974--81\,987, 2022.

\bibitem{valgren2010sift}
C.~Valgren and A.~J. Lilienthal, ``Sift, surf \& seasons: Appearance-based
  long-term localization in outdoor environments,'' \emph{Robotics and
  Autonomous Systems}, vol.~58, no.~2, pp. 149--156, 2010.

\bibitem{arandjelovic2016netvlad}
R.~Arandjelovic, P.~Gronat, A.~Torii, T.~Pajdla, and J.~Sivic, ``Netvlad: Cnn
  architecture for weakly supervised place recognition,'' in \emph{Proceedings
  of the IEEE Conference on Computer Vision and Pattern Recognition}, 2016, pp.
  5297--5307.

\bibitem{Revaud}
J.~Revaud, J.~Almaz{\'a}n, R.~S. Rezende, and C.~R.~d. Souza, ``Learning with
  average precision: Training image retrieval with a listwise loss,'' in
  \emph{IEEE/CVF International Conference on Computer Vision}, 2019.

\bibitem{izquierdo2023optimal}
S.~Izquierdo and J.~Civera, ``Optimal transport aggregation for visual place
  recognition,'' \emph{arXiv preprint arXiv:2311.15937}, 2023.

\bibitem{Nikhil2024Any}
N.~Keetha~\textit{et al.}, ``Anyloc: Towards universal visual place
  recognition,'' \emph{IEEE Robotics and Automation Letters}, vol.~9, no.~2,
  2024.

\bibitem{Ali-bey_2023_WACV}
A.~Ali-bey, B.~Chaib-draa, and P.~Gigu\`ere, ``Mixvpr: Feature mixing for
  visual place recognition,'' in \emph{Proceedings of the IEEE/CVF Winter
  Conference on Applications of Computer Vision}, 2023, pp. 2998--3007.

\bibitem{Hausler2019Multi}
S.~Hausler, A.~Jacobson, and M.~Milford, ``Multi-process fusion: Visual place
  recognition using multiple image processing methods,'' \emph{IEEE Robotics
  and Automation Letters}, vol.~4, no.~2, pp. 1924--1931, 2019.

\bibitem{Malone2023Boosting}
C.~Malone, S.~Hausler, T.~Fischer, and M.~Milford, ``Boosting performance of a
  baseline visual place recognition technique by predicting the maximally
  complementary technique,'' in \emph{2023 IEEE International Conference on
  Robotics and Automation}, 2023, pp. 1919--1925.

\bibitem{pepperell2014all}
E.~Pepperell, P.~I. Corke, and M.~J. Milford, ``All-environment visual place
  recognition with smart,'' in \emph{2014 IEEE international conference on
  robotics and automation (ICRA)}.\hskip 1em plus 0.5em minus 0.4em\relax IEEE,
  2014, pp. 1612--1618.

\bibitem{bai2018sequence}
D.~Bai, C.~Wang, B.~Zhang, X.~Yi, and X.~Yang, ``Sequence searching with cnn
  features for robust and fast visual place recognition,'' \emph{Computers \&
  Graphics}, vol.~70, pp. 270--280, 2018.

\bibitem{garg2021seqnet}
S.~Garg and M.~Milford, ``Seqnet: Learning descriptors for sequence-based
  hierarchical place recognition,'' \emph{IEEE Robotics and Automation
  Letters}, vol.~6, no.~3, pp. 4305--4312, 2021.

\bibitem{bruce2017look}
J.~Bruce, A.~Jacobson, and M.~Milford, ``Look no further: Adapting the
  localization sensory window to the temporal characteristics of the
  environment,'' \emph{IEEE Robotics and Automation Letters}, vol.~2, no.~4,
  pp. 2209--2216, 2017.

\bibitem{kazerouni2022survey}
I.~A. Kazerouni, L.~Fitzgerald, G.~Dooly, and D.~Toal, ``A survey of
  state-of-the-art on visual slam,'' \emph{Expert Systems with Applications},
  vol. 205, p. 117734, 2022.

\bibitem{maddern2012cat}
W.~Maddern, M.~Milford, and G.~Wyeth, ``Cat-slam: probabilistic localisation
  and mapping using a continuous appearance-based trajectory,'' \emph{The
  International Journal of Robotics Research}, vol.~31, no.~4, 2012.

\bibitem{yi2021integrating}
S.~Yi, S.~Worrall, and E.~Nebot, ``Integrating vision, lidar and gps
  localization in a behavior tree framework for urban autonomous driving,'' in
  \emph{2021 IEEE International Intelligent Transportation Systems Conference
  (ITSC)}.\hskip 1em plus 0.5em minus 0.4em\relax IEEE, 2021, pp. 3774--3780.

\bibitem{zhang2018real}
Z.~Zhang, H.~Wang, and W.~Chen, ``A real-time visual-inertial mapping and
  localization method by fusing unstable gps,'' in \emph{13th World Congress on
  Intelligent Control and Automation}, 2018, pp. 1397--1402.

\bibitem{rogers2014mapping}
J.~G. Rogers, J.~R. Fink, and E.~A. Stump, ``Mapping with a ground robot in gps
  denied and degraded environments,'' in \emph{2014 American Control
  Conference}.\hskip 1em plus 0.5em minus 0.4em\relax IEEE, 2014, pp.
  1880--1885.

\bibitem{liu2021visual}
D.~Liu, Y.~Cui, X.~Guo, W.~Ding, B.~Yang, and Y.~Chen, ``Visual localization
  for autonomous driving: Mapping the accurate location in the city maze,'' in
  \emph{International Conference on Pattern Recognition (ICPR)}.\hskip 1em plus
  0.5em minus 0.4em\relax IEEE, 2021.

\bibitem{sarlin2023orienternet}
P.-E. Sarlin~\textit{et al.}, ``Orienternet: Visual localization in 2d public
  maps with neural matching,'' in \emph{Proceedings of the IEEE/CVF Conference
  on Computer Vision and Pattern Recognition}, 2023, pp. 21\,632--21\,642.

\bibitem{chen2014convolutional}
Z.~Chen, O.~Lam, A.~Jacobson, and M.~Milford, ``Convolutional neural
  network-based place recognition,'' in \emph{Australasian Conference on
  Robotics and Automation}, vol.~2, 2014, p.~4.

\bibitem{sunderhauf2015performance}
N.~S{\"u}nderhauf, S.~Shirazi, F.~Dayoub, B.~Upcroft, and M.~Milford, ``On the
  performance of convnet features for place recognition,'' in \emph{Intelligent
  Robots and Systems (IROS), 2015 IEEE/RSJ International Conference on}.\hskip
  1em plus 0.5em minus 0.4em\relax IEEE, 2015, pp. 4297--4304.

\bibitem{garg2018dont}
S.~Garg, N.~Suenderhauf, and M.~Milford, ``Don't look back: Robustifying place
  categorization for viewpoint- and condition-invariant place recognition,'' in
  \emph{IEEE International Conference on Robotics and Automation (ICRA)}, 2018.

\bibitem{malone2022improving}
C.~Malone, S.~Garg, M.~Xu, T.~Peynot, and M.~Milford, ``Improving road
  segmentation in challenging domains using similar place priors,'' \emph{IEEE
  Robotics and Automation Letters}, vol.~7, no.~2, 2022.

\bibitem{vinh2009information}
N.~X. Vinh, J.~Epps, and J.~Bailey, ``Information theoretic measures for
  clusterings comparison: is a correction for chance necessary?'' in
  \emph{Proceedings of the 26th annual international conference on machine
  learning}, 2009, pp. 1073--1080.

\bibitem{chen2017deep}
Z.~Chen~\textit{et al.}, ``Deep learning features at scale for visual place
  recognition,'' in \emph{Robotics and Automation (ICRA), 2017 IEEE
  International Conference on}.\hskip 1em plus 0.5em minus 0.4em\relax IEEE,
  2017, pp. 3223--3230.

\bibitem{RobotCarDatasetIJRR}
W.~Maddern, G.~Pascoe, C.~Linegar, and P.~Newman, ``{1 Year, 1000km: The Oxford
  RobotCar Dataset},'' \emph{The International Journal of Robotics Research},
  vol.~36, no.~1, pp. 3--15, 2017.

\bibitem{Sunderhauf2013}
N.~S{\"u}nderhauf, P.~Neubert, and P.~Protzel, ``Are we there yet? challenging
  seqslam on a 3000 km journey across all four seasons,'' in \emph{Workshop on
  Long-Term Autonomy, IEEE International Conference on Robotics and
  Automation}, 2013.

\bibitem{agarwal2020ford}
S.~Agarwal, A.~Vora, G.~Pandey, W.~Williams, H.~Kourous, and J.~McBride, ``Ford
  multi-av seasonal dataset,'' \emph{The International Journal of Robotics
  Research}, vol.~39, no.~12, pp. 1367--1376, 2020.

\end{thebibliography}

\end{document}